%% file: neurips_2020.tex
\documentclass{article}

\PassOptionsToPackage{numbers, compress}{natbib}
     \usepackage[final]{neurips_2020}

\usepackage[utf8]{inputenc} 
\usepackage[T1]{fontenc}    
\usepackage{hyperref}       
\usepackage{url}            
\usepackage{booktabs}       
\usepackage{amsfonts}       
\usepackage{nicefrac}       
\usepackage{microtype}      

\usepackage{graphicx}
\usepackage{subcaption}
\usepackage{amsmath}
\usepackage{float}
\usepackage{booktabs,multirow}
\usepackage{caption}
\usepackage{subcaption}
\usepackage{color}
\usepackage{url}
\usepackage{multirow}
\usepackage{multicol}
\usepackage[]{todonotes}
\usepackage[ruled, vlined]{algorithm2e}
\usepackage{algorithmic}
\usepackage{rotating}
\usepackage{adjustbox}

\usepackage[capitalise,nameinlink]{cleveref}
 
\crefname{section}{Sec.}{Secs.}
\crefname{appendix}{App.}{Apps.}
\crefname{algorithm}{Alg.}{Algs.}
\creflabelformat{equation}{#2\textup{#1}#3}

\input{defns.tex}

\input{math_commands.tex}

\title{Continual Deep Learning by Functional Regularisation of Memorable Past}

\author{
Pingbo Pan,$^{1,*,\dagger}$ Siddharth Swaroop,$^{2,*,\ddagger}$ Alexander Immer,$^{3,\dagger}$ Runa Eschenhagen,$^{4,\dagger}$ \\
\bf{Richard E. Turner,$^{2}$ Mohammad Emtiyaz Khan$^{5,\ddagger}$}. \\\\
$^1$ University of Technology Sydney, Australia\\
$^2$ University of Cambridge, Cambridge, UK \\
$^3$ École Polytechnique Fédérale de Lausanne, Switzerland\\
$^4$ University of T\"ubingen, T\"ubingen, Germany\\
$^5$ RIKEN Center for AI Project, Tokyo, Japan
}

\begin{document}

\maketitle
\let\svthefootnote\thefootnote
\let\thefootnote\relax\footnotetext{* These two authors contributed equally.}
\let\thefootnote\relax\footnotetext{$\dagger$ This work is conducted during an internship at RIKEN Center for AI project, Tokyo, Japan.}
\let\thefootnote\relax\footnotetext{$\ddagger$ Corresponding authors: \tt{emtiyaz.khan@riken.jp}, \tt{ss2163@cam.ac.uk}}
\addtocounter{footnote}{0}\let\thefootnote\svthefootnote

\begin{abstract}
Continually learning new skills is important for intelligent systems, yet standard deep learning methods suffer from catastrophic forgetting of the past.~Recent works address this with weight regularisation.~Functional regularisation, although computationally expensive, is expected to perform better, but rarely does so in practice.
In this paper, we fix this issue by using a new functional-regularisation approach that utilises a few \emph{memorable past} examples crucial to avoid forgetting.
By using a Gaussian Process formulation of deep networks, our approach enables training in weight-space while identifying both the memorable past and a functional prior.
Our method achieves state-of-the-art performance on standard benchmarks and opens a new direction for life-long learning where regularisation and memory-based methods are naturally combined.
\end{abstract}

\section{Introduction}
\input{sections/1introduction.tex}

\section{Continual Learning with Weight/Functional Regularisation}
\input{sections/2previous_approaches.tex}

\section{Functional-Regularisation of Memorable Past (FROMP)}
\input{sections/3FROMP.tex}

\section{Experiments}\label{sec:experiments}
\input{sections/4experiments.tex}

\section{Discussion}
\input{sections/5discussion.tex}

\newpage
\section*{Broader Impact}
The focus of this paper is on continual deep learning which is related to the field of \emph{life-long learning}.
Designing such algorithms is a bottleneck for deep learning which heavily relies on the \emph{offline} setting where all the data is available at once.
Life-long learning methods, such as ours, will extend the application of deep learning to problems where data is limited and needs to be collected slowly over time.
This could bring a positive change in fields such as robotics, medicine, healthcare, and climate science.
A shortcoming currently is the lack of theoretical guarantees, which is essential to ensure a positive change.
Life-long learning methods, such as ours, should not be applied to mission-critical problems, until such guarantees are available.

One could imagine negative outcomes too, e.g., if life-long learning methods are perfected, machines could then learn in a sequential fashion, similar to living beings and humans. It is possible that their learning will catch up with ours, which will have a huge affect on the society and economy. We do not see this happening any time soon, and in the short term we see a net positive effect on the society. It is important to perform research to understand effects on society in case life-long learning methods are successful.

\section*{Acknowledgements}
Pingbo Pan would like to thank Prof Yi Yang (The ReLER Lab, University of Technology, Sydney) for helpful discussions. 
Runa Eschenhagen is thankful to Rio Yokota for hosting him in Tokyo Institute of Technology during this work.
We are also thankful for the RAIDEN computing system and its support team at the RIKEN Center for AI Project, which we used extensively for our experiments. 
Siddharth Swaroop is supported by an EPSRC DTP studentship.
Richard E.~Turner is supported by Google, Amazon, ARM, Improbable, EPSRC grants EP/M0269571 and EP/L000776/1, and the UKRI Centre for Doctoral Training in the Application of Artificial Intelligence to the study of Environmental Risks (AI4ER).
Mohammad Emtiyaz Khan is partially supported by KAKENHI Grant-in-Aid for Scientific Research (B), Research Project Number 20H04247.

\bibliographystyle{plainnat}
\bibliography{references}

\newpage
\clearpage
\onecolumn
\appendix

\input{sections/appendix/app_A.tex}

\input{sections/appendix/derivation_multiclass.tex}

\input{sections/appendix/functional_approx.tex}
\input{sections/appendix/metrics_app.tex}
\input{sections/appendix/hyperparameters}
\input{sections/appendix/toydata_app.tex}
\input{sections/appendix/task_boundary_detection.tex}
\input{sections/appendix/camera_ready_changes.tex}
\input{sections/appendix/author_contri.tex}

\end{document}

%% file: defns.tex
\newcommand{\loss}{\ell}

\newcommand{\vparam}{\vw}
\newcommand{\param}{w}

\newcommand\cut[1]{}

\newcommand{\squishlist}{
   \begin{list}{$\bullet$}
    { \setlength{\itemsep}{0pt}      \setlength{\parsep}{3pt}
      \setlength{\topsep}{3pt}       \setlength{\partopsep}{0pt}
      \setlength{\leftmargin}{1.5em} \setlength{\labelwidth}{1em}
      \setlength{\labelsep}{0.5em} } }

\newcommand{\squishlisttwo}{
   \begin{list}{$\bullet$}
    { \setlength{\itemsep}{0pt}    \setlength{\parsep}{0pt}
      \setlength{\topsep}{0pt}     \setlength{\partopsep}{0pt}
      \setlength{\leftmargin}{2em} \setlength{\labelwidth}{1.5em}
      \setlength{\labelsep}{0.5em} } }

\newcommand{\squishend}{
    \end{list}  }

{}
{}
{}
{}

{}

{}

\newcommand{\half}{\mbox{$\frac{1}{2}$}}

\newcommand{\real}{\mbox{$\mathbb{R}$}}

\newcommand{\la}{\mbox{$\leftarrow$}}

\newcommand{\rnd}[1]{\left(#1\right)}
\newcommand{\sqr}[1]{\left[#1\right]}

\newcommand{\myexpect}{\mathbb{E}}

\newcommand{\gauss}{\mbox{${\cal N}$}}

\newcommand{\mysoftmax}{\calS}

\newcommand{\myvec}[1]{\mbox{$\mathbf{#1}$}}
\newcommand{\myvecsym}[1]{\mbox{$\boldsymbol{#1}$}}

\newcommand{\vone}{\mbox{$\myvecsym{1}$}}

\newcommand{\vepsilon}{\mbox{$\myvecsym{\epsilon}$}}

\newcommand{\vmu}{\mbox{$\myvecsym{\mu}$}}

\newcommand{\vLambda}{\mbox{$\myvecsym{\Lambda}$}}

\newcommand{\vphi}{\mbox{$\myvecsym{\phi}$}}
\newcommand{\vPhi}{\mbox{$\myvecsym{\Phi}$}}

\newcommand{\vSigma}{\mbox{$\myvecsym{\Sigma}$}}

\newcommand{\va}{\mbox{$\myvec{a}$}}

\newcommand{\vf}{\mbox{$\myvec{f}$}}
\newcommand{\vg}{\mbox{$\myvec{g}$}}
\newcommand{\vh}{\mbox{$\myvec{h}$}}

\newcommand{\vm}{\mbox{$\myvec{m}$}}

\newcommand{\vs}{\mbox{$\myvec{s}$}}

\newcommand{\vu}{\mbox{$\myvec{u}$}}
\newcommand{\vv}{\mbox{$\myvec{v}$}}
\newcommand{\vw}{\mbox{$\myvec{w}$}}

\newcommand{\vx}{\mbox{$\myvec{x}$}}

\newcommand{\vy}{\mbox{$\myvec{y}$}}

\newcommand{\vA}{\mbox{$\myvec{A}$}}

\newcommand{\vF}{\mbox{$\myvec{F}$}}

\newcommand{\vH}{\mbox{$\myvec{H}$}}
\newcommand{\vI}{\mbox{$\myvec{I}$}}
\newcommand{\vJ}{\mbox{$\myvec{J}$}}
\newcommand{\vK}{\mbox{$\myvec{K}$}}

\newcommand{\vX}{\mbox{$\myvec{X}$}}

\newcommand{\diag}{\mbox{$\mbox{diag}$}}

\newcommand{\trace}{\mbox{Tr}}

\newcommand{\calD}{\mbox{${\cal D}$}}

\newcommand{\calS}{\mbox{${\cal S}$}}

\newcommand{\data}{\calD}

\newcommand{\sigmoid}{\mbox{$\sigma$}}

\newcommand{\be}{\begin{equation}}
\newcommand{\ee}{\end{equation}}
\newcommand{\bea}{\begin{eqnarray}}
\newcommand{\eea}{\end{eqnarray}}
\newcommand{\beaa}{\begin{eqnarray*}}
\newcommand{\eeaa}{\end{eqnarray*}}

%% file: math_commands.tex
\usepackage{amsmath,amsfonts,bm}

\newcommand{\memory}{\mathcal{M}}

\def\1{\bm{1}}

\def\rf{{\textnormal{f}}}

\def\ry{{\textnormal{y}}}

\DeclareMathAlphabet{\mathsfit}{\encodingdefault}{\sfdefault}{m}{sl}
\SetMathAlphabet{\mathsfit}{bold}{\encodingdefault}{\sfdefault}{bx}{n}



%% file: sections/1introduction.tex
\label{sec:intro}

The ability to quickly adapt to changing environments is an important quality of intelligent systems.
For such quick adaptation, it is important to be able to identify, memorise, and recall useful past experiences when acquiring new ones.
Unfortunately, standard deep-learning methods lack such qualities, and can quickly forget previously acquired skills when learning new ones \citep{ewc}. 
Such catastrophic forgetting presents a big challenge for applications such as robotics, where new tasks can appear during training, and data from previous tasks might be unavailable for retraining.

In recent years, many methods have been proposed to address catastrophic forgetting in deep neural networks (DNNs).
One popular approach is to keep network weights close to the values obtained for the previous tasks/data \citep{ebrahimi2019uncertainty, ewc, vcl, si}. 
However, this may not always ensure the quality of predictions on previous tasks. 
Since the network outputs depend on the weights in a complex way, such \emph{weight-regularisation} may not be effective.   
A better approach is to use \emph{functional-regularisation}, where we directly regularise the network outputs~\citep{benjamin2018measuring}, but this is costly because it requires derivatives of outputs at many input locations.~Existing approaches reduce these costs by carefully selecting the locations, e.g. by using a \emph{working memory}~\citep{benjamin2018measuring} or Gaussian-Process (GP) inducing points~\citep{titsias2019functional}, but currently they do not consistently outperform existing weight-regularisation methods.

\begin{figure*}[t!]
 \centering
  \begin{subfigure}[b]{0.62\textwidth}
     \centering
     \includegraphics[width=\textwidth]{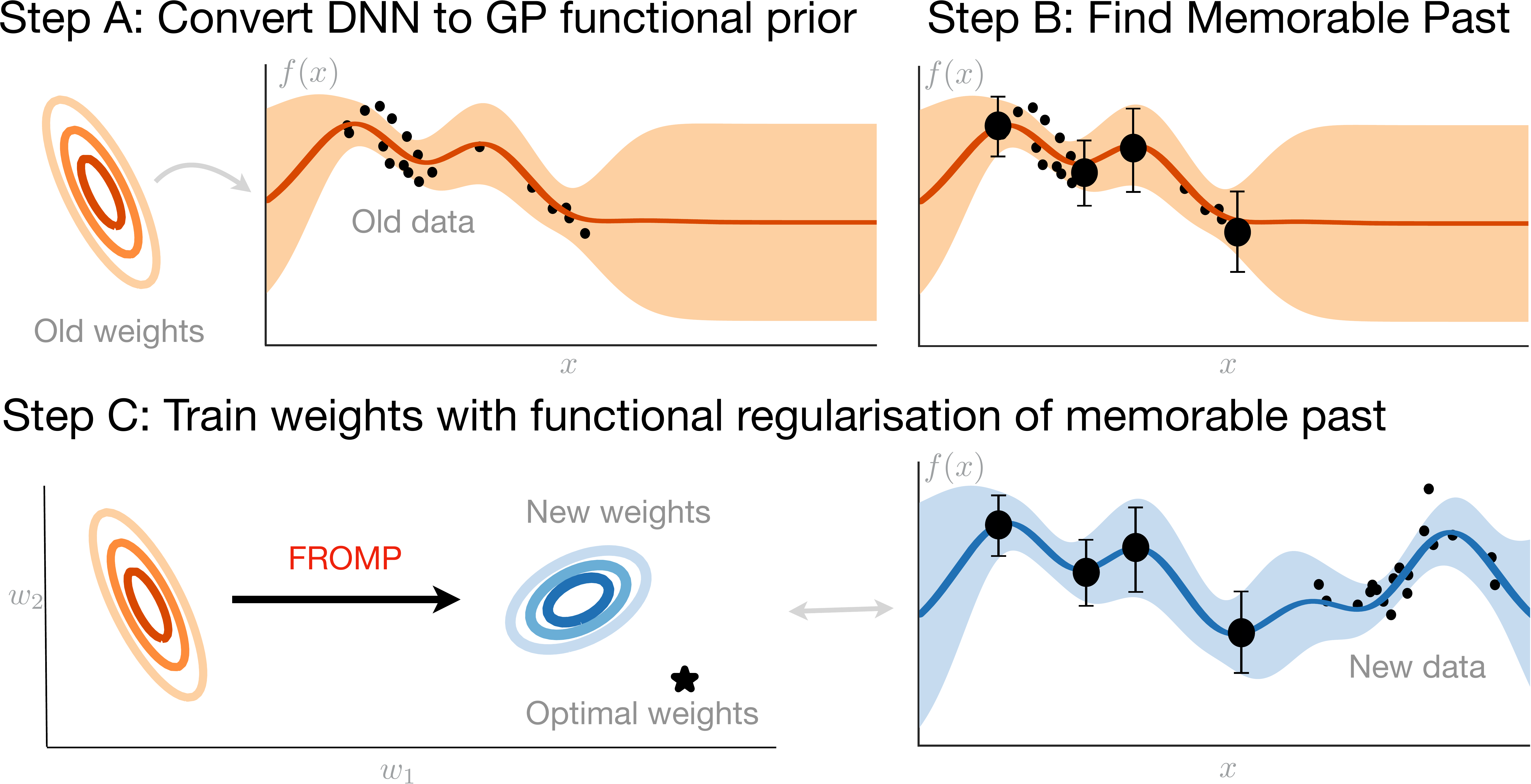}
     \caption{FROMP for continual deep learning}
     \label{fig:cartoon}
 \end{subfigure}
 \hfill
 \begin{subfigure}[b]{0.36\textwidth}
     \centering
     \includegraphics[width=0.465\textwidth]{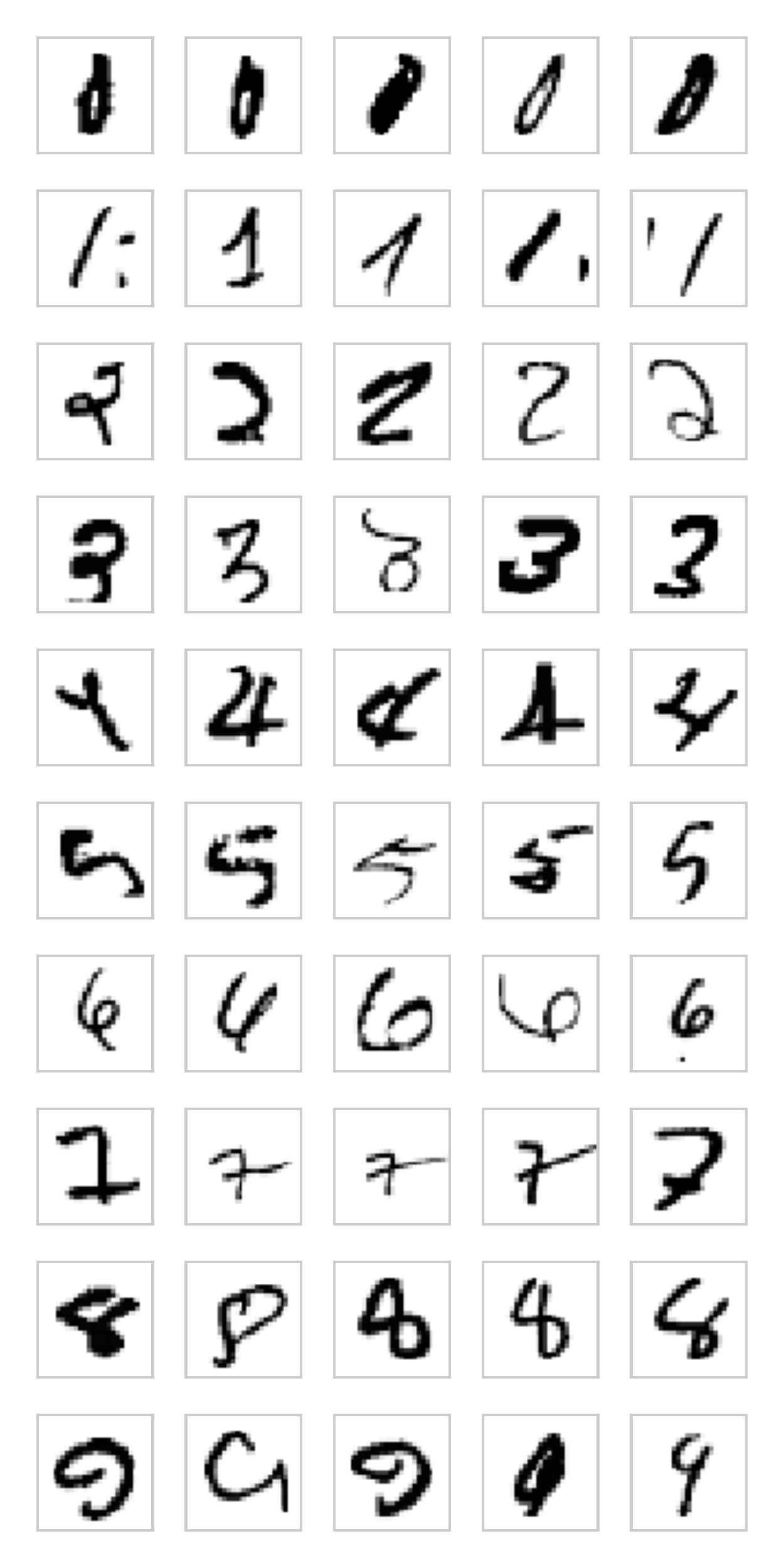}
     \hfill
     \includegraphics[width=0.465\textwidth]{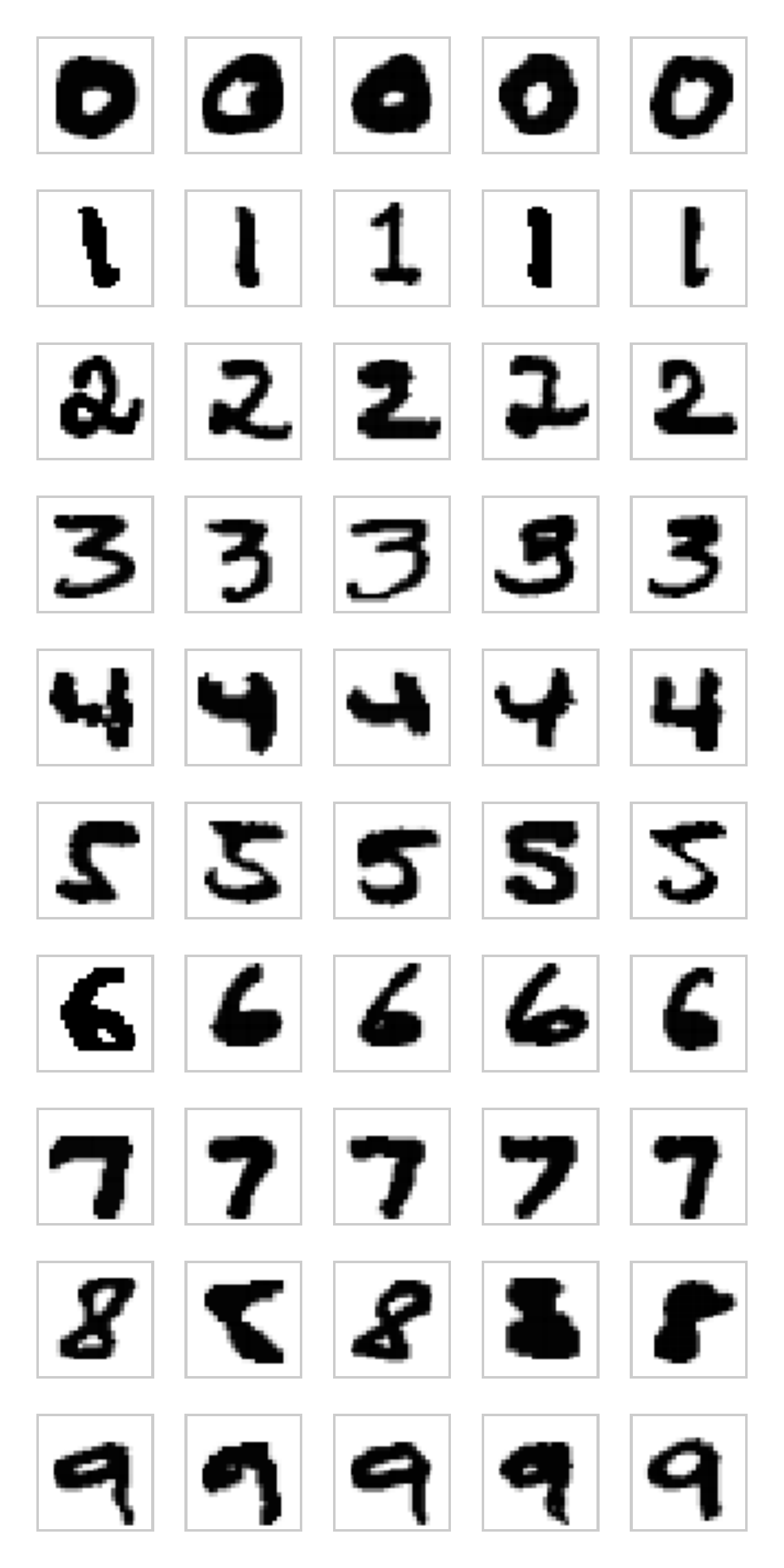}
     \caption{Most (left) vs least (right) memorable }
     \label{fig:memorable_past_mnist}
 \end{subfigure}
    \caption{(a) Our FROMP method consists of three main steps where we convert a DNN to GP using \citet{nn2gp}, find memorable examples, and train weights with functional regularisation of those examples. 
    (b) Memorable past on MNIST -- they are difficult to classify and close to the boundary.  
    }
    \label{fig:cartoon_and_mnist}
\end{figure*}

To address this issue, we propose a new functional-regularisation method called Functional Regularisation of Memorable Past (FROMP).
Our key idea is to regularise the network outputs at a few \emph{memorable past} examples that are crucial to avoid forgetting.
We use a GP formulation of DNNs to obtain a weight-training method that exploits correlations among memorable examples in the function space (see \cref{fig:cartoon}).
FROMP involves a slight modification of Adam and a minor increase in computation cost.
It achieves state-of-the-art performance on standard benchmarks, and is consistently better than both the existing weight-regularisation and functional-regularisation methods. 
Our work in this paper focuses on avoiding forgetting, but it also opens a new direction for life-long learning methods where regularisation methods are naturally combined with memory-based methods.\footnote{Code for all experiments is available at \url{https://github.com/team-approx-bayes/fromp}.}

\subsection{Related Works}

Broadly, existing work on continual learning can be split into three types of approaches: inference-based, memory/rehearsal-based, and model-based.
There have also been hybrid approaches attempting to combine them.
Inference-based approaches have mostly focused on weight regularisation \citep{aljundi2018memory, chaudhry2018riemannian, ebrahimi2019uncertainty, ewc, vcl, si}, with some recent efforts on functional regularisation \citep{benjamin2018measuring, li2016learning, titsias2019functional}. 
Our work falls in the latter category, but also imposes functional constraints at datapoints, thereby connecting to memory-based approaches.

Our goal is to consistently outperform weight-regularisation which can be inadequate and brittle for continual deep learning (see \cref{fig:VCL_toy_runs} and \cref{app:toydata} for an example).
The proposed method further addresses many issues with existing functional-regularisation methods \citep{benjamin2018measuring, titsias2019functional}.
Arguably the work most closely related to ours is the GP-based method of \citet{titsias2019functional}, but there are several key differences. 
First, our kernel uses \emph{all} the network weights (they use just the last layer) which is important, especially in the early stages of learning when all the weights are changing (see \cref{app:importance_all_layers} for an example).
Second, our functional prior regularises the mean to be close to the past mean, which is lacking in the regulariser of \citet{titsias2019functional} (see the discussion after \cref{eq:function_prior}).
Third, our memorable past examples play a similar role as the inducing inputs, but are much cheaper to obtain (\citet{titsias2019functional} requires solving a discrete optimisation problem), and have an intuitive interpretation (see \cref{fig:memorable_past_mnist}).
Due to these differences, our method outperforms the method of \citet{titsias2019functional}, which, unlike ours, performs worse than the weight-regularisation method of \citet{swaroop2019improving}. 
We also obtain state-of-the-art performance on a larger Split CIFAR benchmark, a comparison missing in \citet{titsias2019functional}.
Our method is also different to \citet{benjamin2018measuring}, which lacks a mechanism to automatically weight past memory and estimate uncertainty.

Our method is based on a set of memorable past examples. 
Many such memory-based approaches exist. These either maintain a memory of past data examples \citep{chaudhry2018riemannian, vcl, rebuffi2017icarl} or train generative models on previous tasks to rehearse pseudo-inputs \citep{Shin2017ContinualLW}.
Recent work \citep{aljundi2019gradient, chaudhry2019ontiny} has focused on improving memory-building methods while combining them with inference-based approaches, building on Gradient-Episodic Memory \citep{Chaudhry2018EfficientLL, lopez2017gradient}.
Compared to these approaches, an advantage of our method is that the memory is obtained within the functional-regularisation framework and does not require solving a separate optimisation problem.
The computation is also straightforward, simply requiring a forward-pass through the network followed by picking the top examples (see \cref{sec:choosing mem}).
Finally, model-based approaches change the model architecture during training \citep{fernando2017pathnet, Rusu2016ProgressiveNN, hat}, and this can be combined with other approaches \citep{Schwarz2018ProgressC}. It is possible to use similar features in our GP-based framework. This is an interesting future direction to be pursued.

%% file: sections/2previous_approaches.tex
\label{sec:previous_approaches}
In deep learning, we minimise loss functions to estimate network weights. For example, in supervised multi-class classification problems, we are given a dataset $\data$ of $N$ input-output pairs with outputs $\vy_i$, a one-hot encoded vector of $K$ classes, and inputs $\vx_i$, a vector of length $D$. 
Our goal is to minimise a loss which takes the following form: $N\bar{\loss}(\vparam) + \delta R(\vw)$, where $\bar{\loss}(\vparam) := \frac{1}{N} \sum_{i=1}^{N} \loss( \vy_i, \vf_\param(\vx_i) )$ with deep neural network $\vf_\param (\vx) \in \real^K$ and its weights $\vparam \in \real^P$, $\loss(\vy,\vf)$ denotes a differentiable loss function (e.g., cross entropy) between an output $\vy$ and the network output $\vf$, $R(\vw)$ is a regularisation function (usually an $L_2$-regulariser $R(\vw):=\vw^\top\vw$), and $\delta>0$ controls the regularisation strength. 
Standard deep-learning approaches rely on an unbiased stochastic gradient of the loss $\bar{\loss}$. This usually requires access to all the data examples for all classes throughout training \citep{bottou2010large}. 
It is this unbiased, minibatch setting where deep-learning excels and achieves state-of-the-art performance.

In reality, we do not always have access to all the data at once, and it is not possible to obtain unbiased stochastic gradients. New classes may appear during training and old classes may never be seen again.
For such settings, vanilla mini-batch stochastic-gradient methods lead to catastrophic forgetting of past information \citep{ewc}. 
Our goal in this paper is to design methods that can avoid, or minimise, such catastrophic forgetting.
We focus on a particular setting where the classification task is divided into several tasks, e.g., a task may consist of a classification problem over a subset of classes.
We assume that the tasks arrive sequentially one after the other, and task boundaries are provided to us.
Once the learning is over, we may never see that task again.
Such continual-learning settings have been considered in other works \citep{ewc, vcl, si} with the goal to avoid forgetting of previous tasks.
We also allow storing some past data, which may not always be possible, e.g., due to privacy constraints.

Recent methods have proposed weight-regularisation as a way to combat catastrophic forgetting.
The main idea is to find the important weights for past tasks, and keep new weights close to them. 
For example, when training on the task $t$ while given weights $\vw_{t-1}$ trained on the past tasks, we can minimise the following loss: $N\bar{\loss}_t(\vparam) + \delta (\vparam - \vparam_{t-1})^\top \vF_{t-1}  (\vparam - \vparam_{t-1})$, where $\bar{\loss}_t(\vparam)$ is the loss defined over all data examples from task $t$ and $\vF_{t-1}$ is a preconditioning matrix that favours the weights relevant to the past tasks more than the rest.
The Elastic-Weight Consolidation (EWC) method \citep{ewc} and \citet{ritter2018online}, for example, use the Fisher information matrix as the pre-conditioner, while variational continual learning (VCL) \citep{vcl} employs the precision matrix of the variational approximation.
To reduce the computational complexity, it is common to use a diagonal matrix.
Such weight-space methods reduce forgetting but do not produce satisfactory results.

The challenge in using weight-regularisation lies in the fact that the exact values of the weights do not really matter due to parametric symmetries \citep{benjamin2018measuring, bishop2006pattern}.
Making current weights closer to the previous ones may not always ensure that the predictions on the past tasks also remain unchanged.
Since the network outputs depend on the weights in a complex way, it is difficult to ensure the effectiveness of \emph{weight-regularisation}.   
A better approach is to directly regularise the outputs, because what ultimately matters is the network output, not the values of the weights.
For example, we can use an $L_2$-regulariser over the function values on data examples from past tasks (e.g., see \cite{benjamin2018measuring}) : 
\begin{align}
    \label{eq:l2}
    \min_\param\, N\bar{\loss}_t(\vparam) + \delta  \sum_{s=1}^{t-1} ( \vf_{t,s} - \vf_{t-1,s})^\top( \vf_{t,s} - \vf_{t-1,s})  ,
\end{align}
where $\vf_{t,s}$ and $\vf_{t-1,s}$ are vectors of function values $f_\param(\vx_i)$ and $f_{\param_{t-1}}(\vx_i)$ respectively for all $i \in \data_s$ with $\data_s$ being the dataset for the task $s$.
Rather than making the weights $\vparam$ similar to $\vparam_{t-1}$, such \emph{functional-regularisation} approaches directly force the function values to be similar.
Because of this, we expect them to perform better.
This is also expected for a Bayesian approach, as posterior approximations in the function-space might be better than those in the weight-space.

Unfortunately, such functional-regularisation is computationally infeasible because it requires us to store all past data and compute function values over them.
This computational issue is typically solved by using a subset of inputs.
\citet{benjamin2018measuring} employ a \emph{working memory} \citep{lopez2017gradient, rebuffi2017icarl} while \citet{titsias2019functional} use the inducing point method based on a Gaussian process framework.~As discussed earlier, such approaches do not consistently perform better than existing weight-regularisation methods.
This could be due to the methods they use to build memory or enforce functional regularisation.
Our goal in this paper is to design a functional-regularisation method that is consistently better than weight-regularisation.~We build upon the method of \citet{nn2gp} to convert deep networks into Gaussian processes, as described next.

%% file: sections/3FROMP.tex
\subsection{From Deep Networks to Functional Priors}\label{sec:dnn2gp}
\citet{nn2gp} propose an approach called DNN2GP to convert deep networks to Gaussian processes (GPs). 
We employ such GPs as functional priors to regularise the next task.
The DNN2GP approach is very similar to the standard weight-space to function-space conversion for linear basis-function models \citep{rasmussen2006gaussian}.
For example, consider a linear regression model on a scalar output $y_i = f_\param(\vx_i) + \epsilon_i$ with a function output $f_\param(\vx_i) := \vphi(\vx_i)^\top\vparam$ using a feature map $\vphi(\vx)$.
Assume Gaussian noise $\gauss(\epsilon_i| 0, \Lambda^{-1})$ and a Gaussian prior $\gauss(\vparam|0, \delta^{-1}\vI_P)$ where $\vI_P$ is the identity matrix of size $P\times P$.
It can then be shown that the posterior distribution of this linear model, denoted by $\gauss(\vparam|\vparam_\textrm{lin}, \vSigma_{\textrm{lin}})$, induces a GP posterior on function $f_\param(\vx)$ whose mean and covariance functions are given as follows (see \cref{app:GP_predictive} or  Chapter~2 in \citet{rasmussen2006gaussian}):
\begin{align}
    m_\textrm{lin}(\vx) := f_{\param_\textrm{lin}}(\vx) , \quad  
    \kappa_\textrm{lin}(\vx,\vx') := \vphi(\vx)^\top\, \vSigma_\textrm{lin}\, \vphi(\vx'), \label{eq:lin_gp}
\end{align}
where $\vparam_\textrm{lin}$ is simply the Maximum A Posteriori (MAP) estimate of the linear model, and 
\begin{align}
    \vSigma_{\textrm{lin}}^{-1} := \sum_{i=1}^N \vphi(\vx_i)\, \Lambda\, \vphi(\vx_i)^\top + \delta \vI_P.
    \label{eq:sig_lin}
\end{align}
DNN2GP computes a similar GP posterior but for a \emph{neural network} whose posterior is approximated by a Gaussian. 
Specifically, given a local minimum $\vparam_*$ of the loss $N\bar{\loss}(\vparam) + \frac{1}{2}\delta\vparam^\top\vparam$ for a scalar output $f_\param(\vx)$, we can construct a Gaussian posterior approximation. Following~\citet{nn2gp}, we employ a variant of the Laplace approximation with mean $\vmu_* = \vparam_*$ and covariance
\begin{align}
    \vSigma_*^{-1} = \sum_{i=1}^N \vJ_{\param_*}(\vx_i)^\top\, \Lambda_{\param_*} (\vx_i, \ry_i)\, \vJ_{\param_*}(\vx_i) + \delta \vI_P ,
    \label{eq:cov_dnn2gp}
\end{align}
where $\Lambda_{\param_*}(\vx, \ry):=\nabla_{\rf\rf}^2 \loss(\ry,\rf)$ is the scalar Hessian of the loss function, and $\vJ_{\param_*}(\vx):=\nabla_\param\rf_\param(\vx)^\top$ is the $1\times P$ Jacobian; all quantities evaluated at $\vparam=\vparam_*$. Essentially, this variant uses a Gauss-Newton approximation for the covariance instead of the Hessian.
Comparing \cref{eq:sig_lin,eq:cov_dnn2gp}, we can interpret $\vSigma_*$ as the covariance of a linear model with a feature map $\vphi(\vx) = \vJ_{\param_*}(\vx)^\top$ and noise precision $\Lambda = \Lambda_{\param_*}(\vx,y)$.
Using this similarity, \citet{nn2gp} derive a GP posterior approximation for neural networks.
They show this for a generic loss function (see App. B2 in their paper), e.g., for a regression loss, the mean and covariance functions of the GP posterior take the following form:
\begin{align}
    m_{\param_*}(\vx) &:= f_{\param_*}(\vx), \quad
    \kappa_{\param_*}(\vx,\vx') := \vJ_{\param_*}(\vx)\, \vSigma_*\, \vJ_{\param_*}(\vx')^\top .  \label{eq:laplace_kernel_post_gp}
\end{align}
A similar equation holds for other loss functions such as those used for binary and multiclass classification; see \cref{app:ste} for details.
We denote such GP posteriors by $\mathcal{GP}\rnd{m_{\param_*}(\vx), \kappa_{\param_*}(\vx,\vx')}$, and use them as a \emph{functional prior} to regularise the next task.

The above result holds at a minimiser $\vparam_*$, but can be extended to a sequence of weights obtained during optimisation \cite{nn2gp}.
For example, for Gaussian variational approximations $q(\vparam)$, we can obtain GP posteriors by replacing $\vparam_*$ by a sample $\vparam\sim q(\vparam)$ in \cref{eq:laplace_kernel_post_gp}. 
We denote such GPs by $\mathcal{GP}\rnd{m_{\param}(\vx), \kappa_{\param}(\vx,\vx')}$. 
The result also applies to variants of Newton's method, RMSprop, and Adam (see \cref{app:gp_predictive_von}).
As shown in \citep{nn2gp}, many DNN2GP posteriors are related to the Neural Tangent Kernel (NTK) \citep{ntk}, e.g., the prior distribution to obtain the posterior in \cref{eq:laplace_kernel_post_gp} corresponds to the NTK of a finite-width network.~A slightly different kernel is obtained when a variational approximation is used. Unlike the method of \citet{titsias2019functional}, the kernel above uses \emph{all} the network weights, and uses the Jacobians instead of the network output or its last layer.

\subsection{Identifying Memorable past}
\label{sec:choosing mem}
To reduce the computation cost of functional regularisation, we identify a few memorable past examples. 
To do so, we exploit a property of linear models.
Consider a linear model where different noise precision $\Lambda_i$ is assigned to each pair $\{\vx_i,y_i\}$. For MAP estimation, the examples with high value of $\Lambda_i$ contribute more, as is clear from the objective:
    $\vparam_{\textrm{MAP}} =\arg\max_{\param}   \sum_{i=1}^N \Lambda_i (y_i - \vphi(\vx_i)^\top\vparam)^2 + \delta\, \vparam^\top\vparam$. 
The noise precision $\Lambda_i$ can therefore be interpreted as the relevance of the data example $i$.
Such relevant examples are crucial to ensure that the solution stays at $\vparam_{\textrm{MAP}}$ or close to it. 
These ideas are widely used in the theory of leverage-score sampling \citep{alaoui2015fast, ma2015statistical} to identify the most \emph{influential} examples.
Computation using such methods is infeasible since they require inverting a large matrix.
\citet{titsias2019functional} use an approximation by inverting smaller matrices, but they require solving a discrete optimisation problem to select examples. 
We propose a method which is not only cheap and effective, but also yields intuitive results.

We use the linear model corresponding to the GP posterior from Section~\ref{sec:dnn2gp}. 
The linear model assigns different noise precision to each data example. See \cref{eq:sig_lin,eq:cov_dnn2gp} where the quantity $\Lambda_{\param_*}(\vx_i,y_i)$ plays the same role as the noise precision $\Lambda$. 
Therefore, $\Lambda_{\param_*}(\vx_i,y_i)$ can be used as a relevance measure,
and a simple approach to pick influential examples is to sort it $\forall i$ and pick the top few examples.
We refer to such a set of examples as the \emph{memorable past} examples.
An example is shown in \cref{fig:memorable_past_mnist} where our approach picks many examples that are difficult to classify. 
The memorable past can be intuitively thought of as \emph{examples close to the decision boundary}.
An advantage of using this approach is that $\Lambda_{\param_*}(\vx_i,y_i)$ is extremely cheap to compute. It is simply the second derivative of the loss, which can be obtained with a forward pass to get $\loss(y_i, \hat{y}_i)$, followed by double differentiation with respect to $\hat{y}_i$. 
For binary classification, our approach is equivalent to the ``Confidence Sampling'' approaches used in the Active Learning literature~\citep{ash2020batch, Wang2014Active}, although in general it differs from them. 
After training on task $t$, we select a set of few memorable examples in $\data_t$, which we denote by $\memory_t$.

\subsection{Training in weight-space with a functional prior}
\label{sec:putting together}
We will now describe the final step for weight-training with functional-regularisation. 
We use the Bayesian formulation of continual learning and replace the prior distribution in weight space by a \emph{functional prior}.
%
Given a loss of the form $N\bar{\loss}_t(\vparam) + R(\vparam)$, a Bayesian formulation in weight-space employs a regulariser that uses the previous posterior, i.e., $R(\vparam) \equiv - \log p(\vparam|\data_{1:t-1})$.
Computing the exact posterior, or a tempered version of it, would in theory avoid catastrophic forgetting, but that is expensive and we must use approximations. For example, \citet{vcl} use the variational approximation from the previous task $p(\vparam | \data_{1:t-1}) \approx q_{t-1}(\vparam) =\gauss(\vparam|\vmu,\vSigma)$ to obtain the weight regulariser. 
Our goal is to replace such weight regulariser by a functional regulariser obtained by using the GP posteriors described in \cref{sec:dnn2gp}.

We use functional regularisation defined over memorable examples. Denote by $\vf$ the vector of function values defined at all memorable past $\mathcal{M}_{s}$ in all tasks $s<t$.
Denoting a sample from $q(\vparam)$ by $\vparam$, we can obtain a GP posterior over $\vf$ by using \cref{eq:laplace_kernel_post_gp}. We denote it by $\tilde{q}_{\param}(\vf) = \gauss(\vf| \vm_t(\vparam), \vK_t(\vparam))$, where $\vm_t(\vparam)$ and $\vK_t(\vparam)$ respectively denote the mean vector and kernel matrix obtained by evaluating $\mathcal{GP}\rnd{m_{\param}(\vx), \kappa_{\param}(\vx,\vx')}$ at the memorable past examples.
Similarly, denoting a sample from $q_{t-1}(\vparam)$ by $\vparam_{t-1}$, we can obtain another GP posterior, which we call the \emph{functional prior}, denoted by $\tilde{q}_{\param_{t-1}}(\vf) = \gauss(\vf|\vm_{t-1}, \vK_{t-1})$. Using these two GPs, we can replace the weight regulariser used in \citep{vcl} by a \emph{functional regulariser} which is equal to the expectation of the functional prior:
\begin{align}
    \min_{q(\param)} \; \myexpect_{q(\param)} &\left[ (N/\tau) \bar{\loss}_t(\vparam) + \log q(\vparam)\right] - \underbrace{ \myexpect_{q(\param)} \left[\log q_{t-1}(\vparam)\right]}_{\approx \, \myexpect_{\tilde{q}_{\param}(\mathbf{f})} \sqr{\log \tilde{q}_{\param_{t-1}}(\mathbf{f})}}  ,
    \label{eq:continual_loss}
\end{align}
where the last term is the weight regulariser, and $\tau>0$ is a tempering parameter. Fortunately, the functional regulariser has a closed-form expression: $\myexpect_{\tilde{q}_{\param}(\rf)} \sqr{\log \tilde{q}_{\param_{t-1}}(\vf)} =$
\begin{align}
     -\half\sqr{\trace{(\vK_{t-1}^{-1} \vK_t(\vparam))} + (\vm_t(\vparam) - \vm_{t-1})^\top \vK_{t-1}^{-1} (\vm_t(\vparam) - \vm_{t-1})} + \textrm{constant}. \label{eq:function_prior}
\end{align}
This term depends on $\vmu$ and $\vSigma$ through the sample $\vparam \sim q(\vparam)$. The regulariser is an approximation for reasons discussed in \cref{app:functional_approx}.
The regulariser has a similar form\footnote{Their regulariser is $\myexpect_{q(u_{t-1})} [\log p_\param(\vu_{t-1})] = -\half\{ \trace[ \vK(\vparam)^{-1} \vSigma_{t-1}] +  \vmu_{t-1}^\top \vK(\vparam)^{-1}\vmu_{t-1} \} $ where $p_\param(\vu_{t-1})= \gauss(0, K(\vparam))$ with the kernel evaluated at inducing inputs $\vu_{t-1}$ and $q(\vu_{t-1}) = \gauss(\vmu_{t-1},\vSigma_{t-1})$. This regulariser encourages $\vK(\vparam)$ to remain close to the second moment $\vSigma_{t-1} + \vmu_{t-1} \vmu_{t-1}^\top$.} to \citet{titsias2019functional}, but unlike their regulariser, ours forces the mean $\vm_t(\vparam)$ to be close to $\vm_{t-1}$, which is desirable since it encourages the predictions of the past to remain the same.

Optimising $\vmu$ and $\vSigma$ in \cref{eq:continual_loss} with this functional prior can be very expensive for large networks. We make five approximations to reduce the cost, discussed in detail in \cref{app:derivation}.
First, for the functional prior, we use the mean of $q_{t-1}(\vparam)$, instead of a sample $\vparam_{t-1}$, which corresponds to using the GP posterior of \cref{eq:laplace_kernel_post_gp}. 
Second, for \cref{eq:function_prior}, we ignore the derivative with respect to $\vK_t(\vparam)$ and only use $\vm_t(\vparam)$, which assumes that the Jacobians do not change significantly.
Third, instead of using the full $\vK_{t-1}$, we factorise it across tasks, i.e., let it be a block-diagonal matrix with $\vK_{t-1,s}, \forall s$ as the diagonal.
This makes the cost of inversion linear in the number of tasks.
Fourth, following \citet{nn2gp}, we propose to use a deterministic optimiser for \cref{eq:continual_loss}, which corresponds to setting $\vparam = \vmu$.
Finally, we use a diagonal $\vSigma$, which corresponds to a mean-field approximation, reducing the cost of inversion.
As shown in \cref{app:derivation}, the resulting algorithm finds a solution to the following problem:
\begin{align}
    \label{eq:fromp_final_reg}
   \min_\param N\bar{\loss}_t(\vparam) + \half \tau \sum_{s=1}^{t-1} \sqr{\vm_{t,s}(\vparam) - \vm_{t-1,s}}^\top \vK_{{t-1},s}^{-1} \sqr{\vm_{t,s}(\vparam) - \vm_{t-1,s}} ,
\end{align}
where $\vm_{t,s}$ is the sub-vector of $\vm_t$ corresponding to the task $s$. 
The above is a computationally-cheap approximation of \cref{eq:continual_loss} and forces the network to produce similar outputs at memorable past examples. 
The objective is an improved version of \cref{eq:l2} \citep{benjamin2018measuring}.
For regression, the mean $\vm_{t,s}$ in \cref{eq:fromp_final_reg} is equal to the vector $\vf_{t,s}$ used in \cref{eq:l2}.
Our functional regulariser additionally includes a kernel matrix $\vK_{t-1,s}$ to take care of the uncertainty and weighting of past tasks' memorable examples. 

Due to a full kernel matrix, the functional regulariser exploits the correlations between memorable examples.~This is in contrast with a weight-space approach, where modelling correlations is infeasible since $\vSigma$ is extremely large. 
Here, training is cheap since the objective in \cref{eq:fromp_final_reg} can be optimised by using Adam. \emph{Our approach therefore provides a cheap weight-space training method while exploiting correlations in function-space.}
Due to these properties, we expect our method to perform better.
We can expect further improvements by relaxing these assumptions (see \cref{app:derivation}), e.g., we can use a full kernel matrix, use a variational approximation, or employ a block-diagonal covariance matrix. We leave such comparisons as future work since they require sophisticated implementation to scale.

\begin{algorithm}[t]
    \caption{FROMP for binary classification on task $t$ given $q_{t-1}(\vparam) := \gauss(\vmu_{t-1}, \diag(\vv_{t-1}))$, and memorable pasts $\memory_{1:t-1}$. Additional computations on top of Adam are highlighted in red.
    }
    \label{alg:fromp}
    \DontPrintSemicolon
    \begin{multicols}{2}
    \SetAlgoLined
    \SetKwFunction{FMain}{FROMP}
    \SetKwProg{Fn}{Function}{:}{}
    \Fn{\FMain{$\data_t, \vmu_{t-1}, \vv_{t-1}, \memory_{1:t-1}$}}{
        \textcolor{red}{Get $\vm_{t-1, s}, \vK^{-1}_{t-1, s}, \forall$ tasks $s<t$} (\cref{eq:mK_bin}) \;
        Initialise $\vparam \leftarrow \vmu_{t-1}$\;
        \While{not converged}{
            Randomly sample $\{\vx_i,y_i\}\in\data_t$\\
            $\vg \leftarrow N\, \nabla_{\param}\loss(y_i, f_\param(\vx_i))$\\
            $\vg_f \leftarrow $ \textcolor{red}{{\tt g\_{FR}} ($ \vparam, \vm_{t-1}, \vK^{-1}_{t-1}, \memory_{1:t-1} $)}\\
            Adam update with gradient $\vg + \tau\vg_f$
        }
        $\vmu_t \leftarrow \vparam$ and {\color{red} compute $\vv_t$} (\cref{eq:cov_bin})\\
        $\memory_t \, \la$ \textcolor{red}{{\tt memorable\_past}$\left(\data_t, \vw \right)$}\\
        \KwRet $\; \vmu_t, \vv_t, \memory_t$\;
    }
    \SetKwFunction{FMain}{g\_FR}
    \SetKwProg{Pn}{Function}{:}{}
    \Pn{\FMain{$\vparam, \vm_{t-1}, \vK^{-1}_{t-1}, \memory_{1:t-1} $}}{
        Initialise $\vg_f \leftarrow\mathbf{0}$\\
        \For{task $s=1,2,...,t-1$}{
        Compute $\vm_{t,s}$ (\cref{eq:mK_bin})\\
        $\vh_i \leftarrow \Lambda_{\param}(\vx_i)\, \vJ_{\param}(\vx_i)^\top, \forall \vx_i\in\memory_s$\\
        Form matrix $\vH$ with $\vh_i$ as columns\\
        $\vg_f  \leftarrow \vg_f  + \vH \vK_{t-1,s}^{-1} (\vm_{t,s}- \vm_{t-1,s})$}
        \KwRet $ \vg_f $\;
    }
    \SetKwFunction{FMain}{memorable\_past}
    \SetKwProg{Pn}{Function}{:}{}
    \Pn{\FMain{$\data_t, \vparam$}}{
        Calculate $\Lambda_{\param}(\vx_i), \,\, \forall \vx_i \in \data_t$.\\
        \KwRet $M$ examples with highest $\Lambda_{\param}(\vx_i)$. 
    }
    \end{multicols}
    \vspace{.2cm}
\end{algorithm}

\subsection{The final algorithm and computational complexity}
\label{sec:final_algo}
The resulting algorithm, FROMP, is shown in \cref{alg:fromp} for binary classification (extension to multiclass classification is in \cref{app:multiclass}).
For binary classification, we assume a sigmoid $\sigmoid(f_\param(\vx))$ function and cross-entropy loss.
As shown in \cref{app:ste}, the Jacobian (of size $1\times P$) and noise precision (a scalar) are as follows:
$\vJ_\param(\vx) = \nabla_\param f_\param(\vx) ^\top$ and $\Lambda_\param(\vx) = \sigma\rnd{f_\param(\vx)} \sqr{1- \sigma\rnd{f_\param(\vx)}}$.
To compute the mean and kernel, we need the diagonal of the covariance, which we denote by $\vv$. This can be obtained using \cref{eq:cov_dnn2gp} but with the sum over $\data_{1:t}$. The update below computes this recursively: 
\begin{align}
    \frac{\vone}{\vv_t} = \Big[ \frac{\vone}{\vv_{t-1}} + \sum_{i\in\data_t}   \diag\rnd{ \vJ_{\param}(\vx_i)^\top \Lambda_{\param}(\vx_i) \vJ_{\param}(\vx_i)} \Big], 
    \label{eq:cov_bin}
\end{align}
where `/' denotes element-wise division and $\diag(\vA)$ is the diagonal of $\vA$.
Using this in an expression similar to \cref{eq:laplace_kernel_post_gp}, we can compute the mean and kernel matrix (see \cref{app:ste} for details):
\begin{align}
    \vm_{t,s}(\vparam)[i] &= \sigma\rnd{f_{\param}(\vx_i)},\,\,  
\vK_{t,s}(\vparam)[i,j] = \Lambda_{\param}(\vx_i) \sqr{\vJ_{\param}(\vx_i)\, \textrm{Diag}\rnd{\vv_t} \vJ_{\param}(\vx_j)^\top} \Lambda_{\param}(\vx_j)\, , 
\label{eq:mK_bin}
\end{align}
over all memorable examples $\vx_i,\vx_j$, where $\textrm{Diag}(\va)$ denotes a diagonal matrix with $\va$ as the diagonal. Using these, we can write the gradient of \cref{eq:fromp_final_reg}, where the gradient of the functional regulariser is added as an additional term to the gradient of the loss:
    $N \nabla_{\param} \bar{\loss}_t(\vparam) + \tau \, \sum_{s=1}^{t-1} (\nabla_{\param}\vm_{t,s}(\vparam)) \vK_{t-1,s}^{-1} (\vm_{t,s}(\vparam) - \vm_{t-1,s})$
where   
    $\nabla_{\param}\vm_{t,s}(\vparam)[i]= \nabla_{\param} \sqr{  \sigma\rnd{f_{\param}(\vx_i)}}=  \Lambda_{\param}(\vx_i) \vJ_{\param}(\vx_i)^\top$.
The regulariser is computed in subroutine {\tt g\_FR} in \cref{alg:fromp}.

The additional computations on top of Adam are highlighted in red in \cref{alg:fromp}.
Every iteration requires functional gradients (in {\tt g\_FR}) whose cost is dominated by the computation of $\vJ_{\param}(\vx_i)$ at all $\vx_i \in \memory_s, \forall s < t$.
Assuming the size of the memorable past is $M$ per task, this adds an additional $\mathcal{O}(MPt)$ computation, where $P$ is the number of parameters and $t$ is the task number. This increases only linearly with the size of the memorable past. 
We need three additional computations but they are required only \emph{once per task}. First,  inversion of $\vK_s, \forall s<t$, which has cost $O(M^3t)$. This is linear in number of tasks and is feasible when $M$ is not too large.
Second, computation of $\vv_t$ in \cref{eq:cov_bin} requires a full pass through the dataset $\data_t$, with cost $O(NP)$ where $N$ is the dataset size. 
This cost can be reduced by estimating $\vv_t$ using a minibatch of data (as is common for EWC \citep{ewc}).
Finally, we find the memorable past $\memory_t$, requiring a forward pass followed by picking the top $M$ examples.

%% file: sections/4experiments.tex
To identify the benefits of the functional prior (step A) and memorable past (step B), we compare FROMP to three variants: (1) FROMP-$L_2$ where we replace the kernel in \cref{eq:laplace_kernel_post_gp} by the identity matrix, similar to \cref{eq:l2}, (2) FRO\emph{R}P where memorable examples selected randomly (``R'' stands for random), (3) FRO\emph{R}P-$L_2$ which is same as FRO\emph{R}P, but the kernel in \cref{eq:laplace_kernel_post_gp} is replaced by the identity matrix.
We present comparisons on four benchmarks: a toy dataset, permuted MNIST, Split MNIST, and Split CIFAR (a split version of CIFAR-10 \& CIFAR-100). Results for the toy dataset are summarised in \cref{fig:app_illustration} and \cref{app:toydata}, where we also visually show the brittleness of weight-space methods. 
In all experiments, we use the Adam optimiser~\citep{adam}. Details on hyperparameter settings are in \cref{app:hypers}. 

\subsection{Permuted and Split MNIST}
Permuted MNIST consists of a series of tasks, with each applying a fixed permutation of pixels to the entire MNIST dataset. Similarly to previous work \citep{ewc, si, vcl, titsias2019functional}, we use a fully connected single-head network with two hidden layers, each consisting of 100 hidden units. We train for 10 tasks. The number of memorable examples is set in the range 10--200. 
%
We also test on the Split MNIST benchmark ~\citep{si}, which consists of five binary classification tasks built from MNIST: 0/1, 2/3, 4/5, 6/7, and 8/9. Following the settings of previous work, we use a fully connected multi-head network with two hidden layers, each with 256 hidden units. We select 40 memorable points per task.

\begin{figure}[t]
    \centering
    \begin{subfigure}[h]{0.56\textwidth}
        \begin{tabular}{lll}
            \toprule 
            \textbf{Method} & \textbf{Permuted} & \textbf{Split}\\
            \hline
             DLP~\citep{smola2003laplace} & $82\%$ & $61.2\%$\\
             EWC~\citep{ewc} & $84\%$ & $63.1\%$ \\
             SI~\citep{si} & $86\%$ & $98.9\%$ \\
             Improved VCL~\citep{swaroop2019improving} & $93 \pm 1\%$ & $98.4 \pm 0.4\%$ \\
              \; + random Coreset & $\mathbf{94.6} \pm 0.3\%$ & $98.2 \pm 0.4\%$\\
             FRCL-RND~\citep{titsias2019functional} & $94.2 \pm 0.1\%$ & $97.1 \pm 0.7\%$ \\
             FRCL-TR~\citep{titsias2019functional} & $94.3 \pm 0.2\%$ & $97.8 \pm 0.7\%$ \\
             \hline
             FRORP-$L_2$ & $87.9 \pm 0.7\%$ & $98.5 \pm 0.2\%$ \\
             FROMP-$L_2$ & $94.6 \pm 0.1\%$ & $98.7 \pm 0.1\%$ \\
             FRORP & $94.6 \pm 0.1\%$ & $\mathbf{99.0} \pm 0.1\%$  \\
             FROMP & $\mathbf{94.9} \pm 0.1\%$ & $\mathbf{99.0} \pm 0.1\%$ \\
            \bottomrule
        \end{tabular}
        \caption{MNIST comparisons: for Permuted, we use 200 examples as memorable/coreset/inducing points. For Split, we use 40.}
        \label{tab:mnist}
    \end{subfigure}
    \hfill
    \begin{subfigure}[h]{0.38\textwidth}
         \centering
         \includegraphics[width=0.49\textwidth]{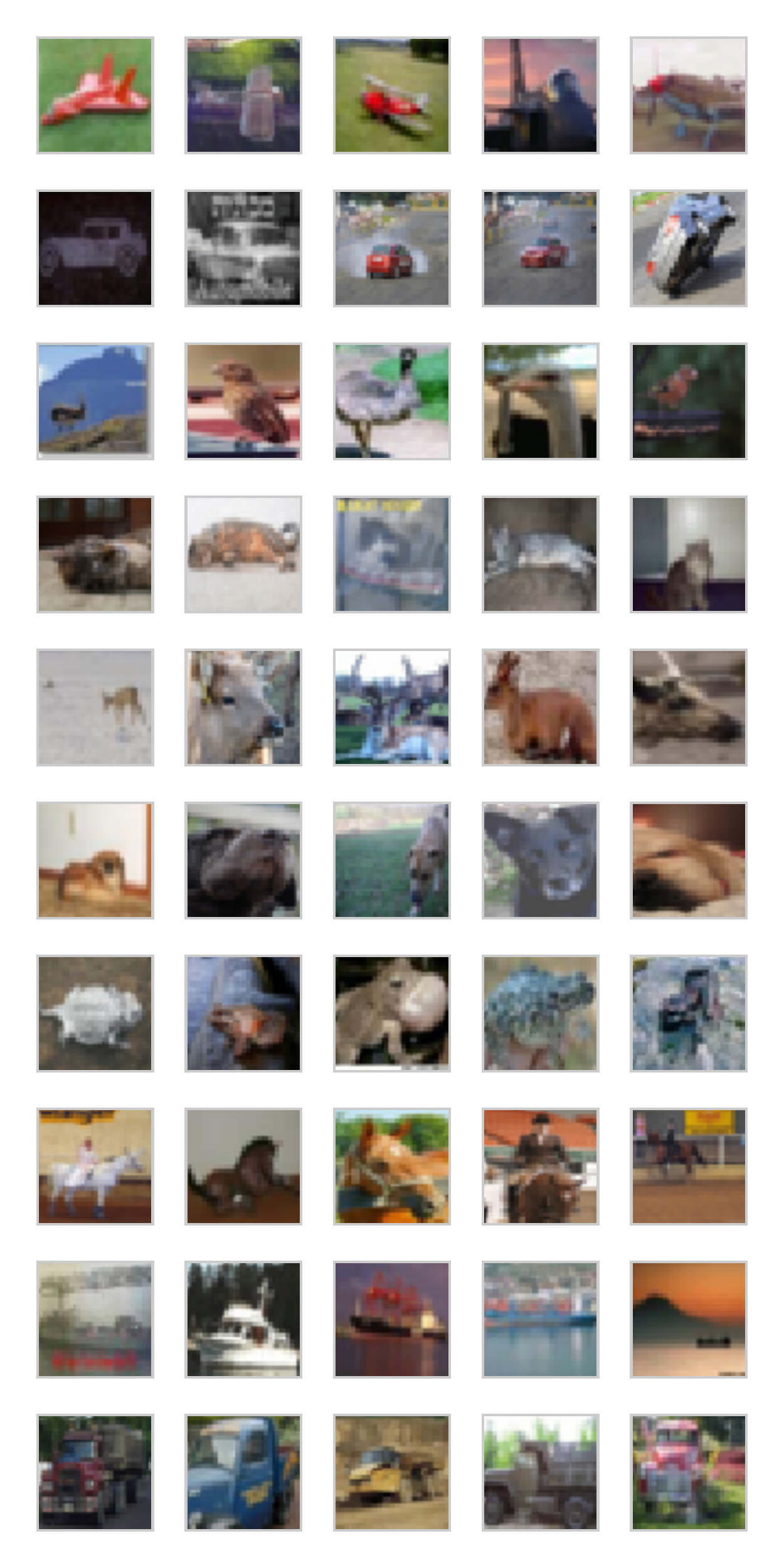}
         \hfill
         \includegraphics[width=0.49\textwidth]{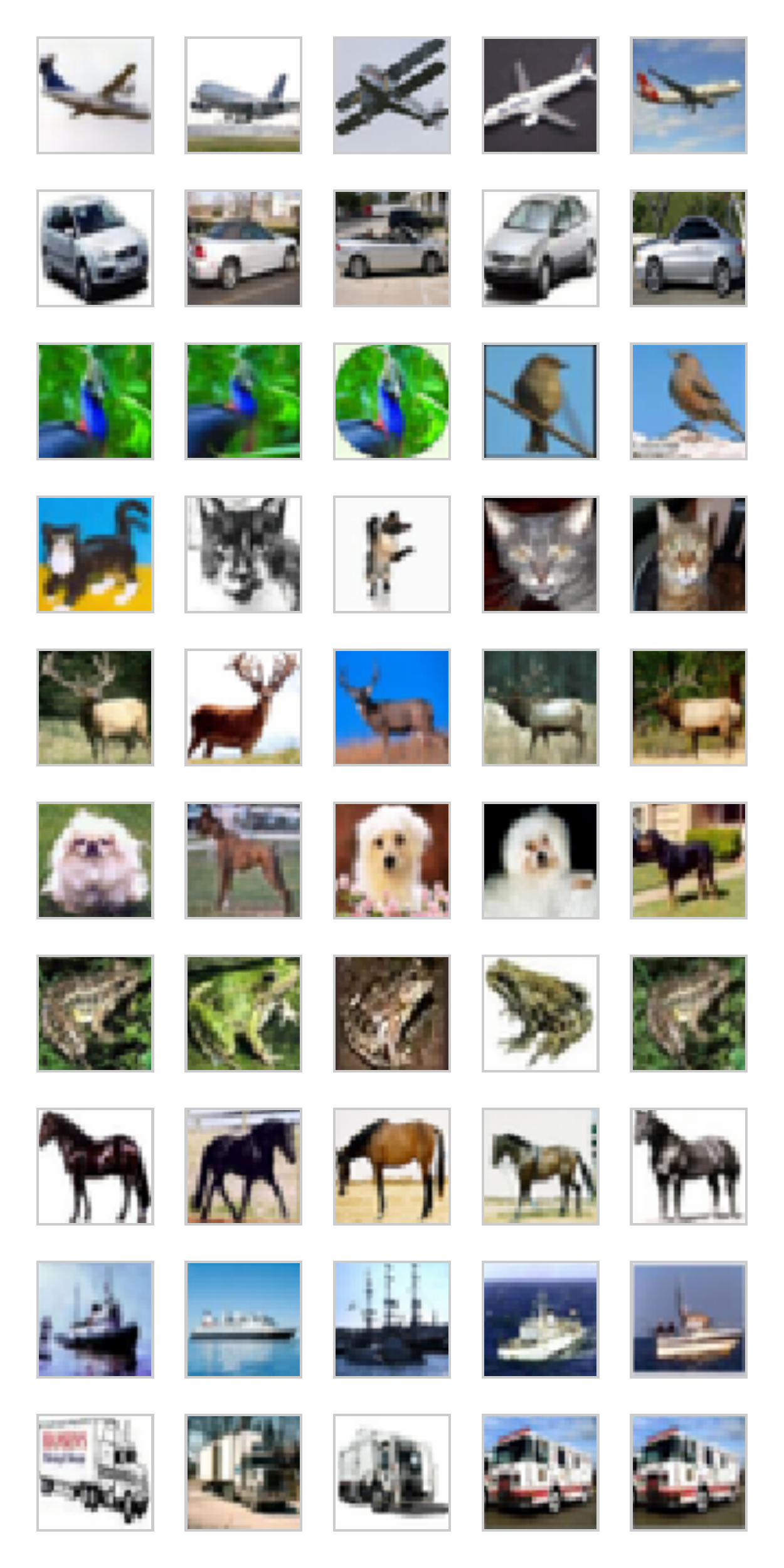}
         \caption{Most (left) vs least (right) memorable}
         \label{fig:memorable_past_cifar}
    \end{subfigure}
    \caption{(a) On MNIST, FROMP obtains better accuracy than weight-regularisation (EWC, SI, VCL) and functional-regularisation (FRCL). Note that FRCL does not outperform `Improved VCL + random coreset' while FROMP does. The standard errors are reported over 5 runs.
    }
    \label{fig:mnist_results_cifar_plot}
\end{figure}

The final average accuracy is shown in \cref{tab:mnist} where FROMP achieves better performance than weight-regularisation methods (EWC, VCL, SI) as well as the function-regularisation continual learning (FRCL) method~\cite{titsias2019functional}.
FROMP also improves over FRORP-$L_2$ and FROMP-$L_2$, demonstrating the effectiveness of the kernel. 
The improvement compared to FRORP is not significant.
We believe this is because a random memorable past already achieves a performance close to the highest achievable performance, and we see no further improvement by choosing the examples carefully.
However, as shown in \cref{fig:permuted_mem_vs_acc}, we do see an improvement when the number of memorable examples are small (compare FRORP vs FROMP). 
Finally, \cref{fig:memorable_past_mnist} shows the most and least memorable examples chosen by sorting $\Lambda_\param(\vx,y)$. 
The most memorable examples appear to be more difficult to classify than the least memorable examples, which suggests that they may lie closer to the decision boundary. 

We also run FROMP on Split MNIST on a smaller network architecture \citep{swaroop2019improving}, obtaining $(99.2\pm0.1)\%$ (see \cref{app:split_mnist}). 
Additionally, in \cref{app:task_boundary_detection}, we show that, when the task-boundary information is unavailable, it is still possible to automatically detect the boundaries within our method. 
When new tasks are encountered, we expect the prediction using current network and past ones to be similar. 
We use this to detect task boundaries by performing a statistical test; see \cref{app:task_boundary_detection} for details.

\subsection{Split CIFAR}
Split CIFAR is a more difficult benchmark than MNIST, and consists of 6 tasks. The first task is the full CIFAR-10 dataset, followed by 5 tasks, each consisting of 10 consecutive classes from CIFAR-100. 
We use the same model architecture as \citet{si}: a multi-head CNN with 4 convolutional layers, then 2 dense layers with dropout. 
The number of memorable examples is set in the range 10--200, and we run each method 5 times. 
We compare to two additional baselines. The first baseline consists of networks trained on each task \emph{separately}. Such training cannot profit from forward/backward transfer from other tasks, and sets a lower limit which we must outperform. 
The second baseline is where we train all tasks \emph{jointly}, which would yield perhaps the best results and which we would like to match. 

\begin{figure}[t]
 \centering
     \begin{subfigure}{3.0in}
         \centering
         \includegraphics[width=3.0in]{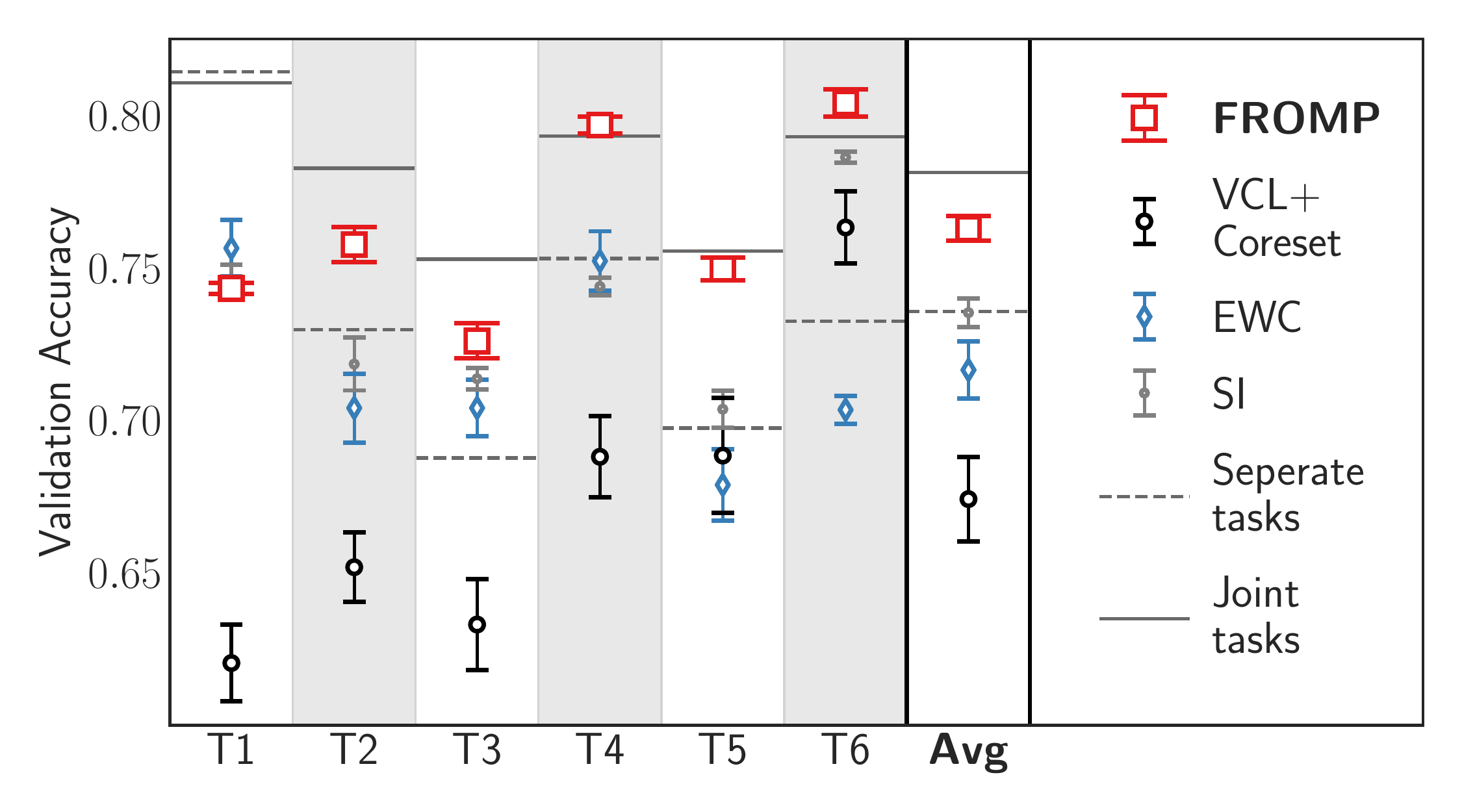}
         \caption{Split CIFAR: Individual task accuracy and their average}
         \label{fig:bar_plot_cifar}
     \end{subfigure}
     \begin{subfigure}{1.2in}
         \centering
         \includegraphics[width=1.2in]{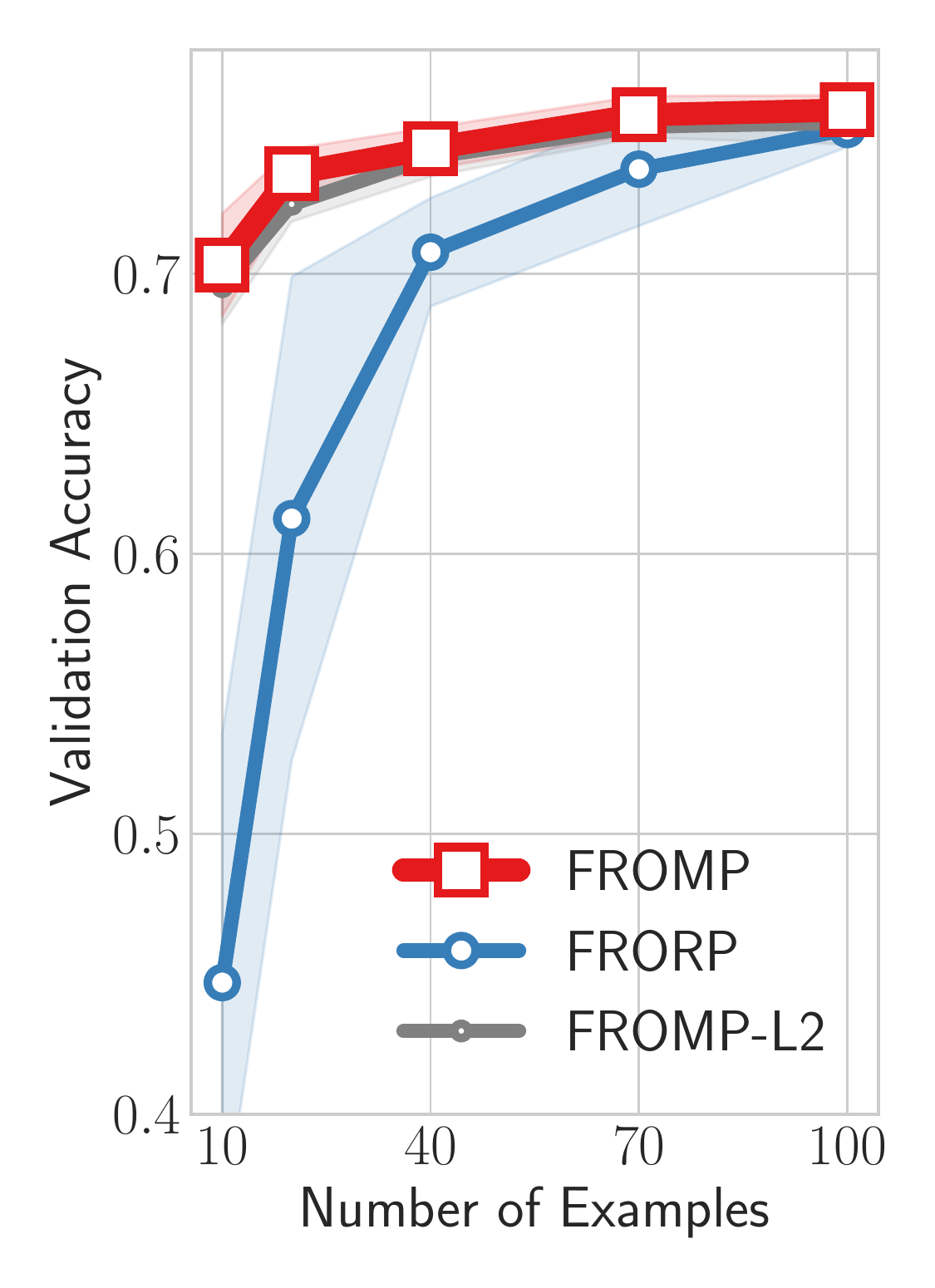}
         \caption{Split CIFAR}
         \label{fig:mem_vs_acc_splitcifar}
     \end{subfigure}
     \begin{subfigure}{1.2in}
        \centering
        \includegraphics[width=1.2in]{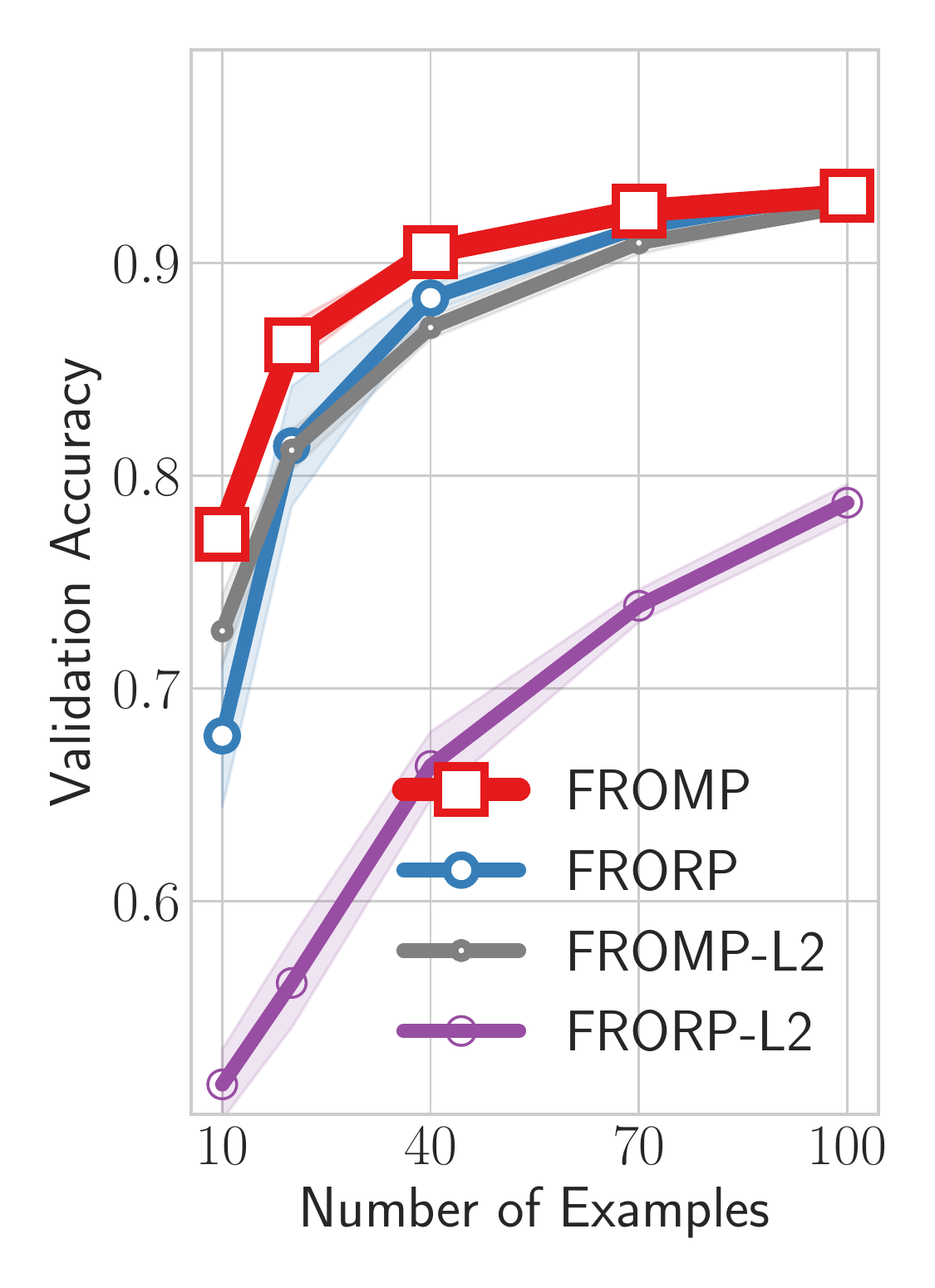}
        \caption{Permuted MNIST}
        \label{fig:permuted_mem_vs_acc}
     \end{subfigure}
    \caption{Fig.~(a) shows that FROMP outperforms weight-regularisation methods  (see \cref{app:split_cifar} for numerical values). `Tx' means Task x. Figs.~(b) and~(c) show average accuracy with respect to the number of memorable examples. A careful selection of memorable examples in FROMP gives better results than random examples in FRORP, especially when the memory size is small. For MNIST, the kernel in FROMP improves performance over FROMP-L2, which does not use a kernel.
}
    \label{fig:past_samples}
\end{figure}

The results are summarised in \cref{fig:bar_plot_cifar}, where we see that FROMP is close to the upper limit while outperforming all the other methods. 
The weight-regularisation methods EWC and SI do not perform well on the later tasks while VCL forgets the earlier tasks. Poor performance of VCL is most likely due to the difficulty of using Bayes By Backprop \citep{blundell2015weight} on CNNs\footnote{Previous results by \citet{vcl} and \citet{swaroop2019improving} are obtained using multi-layer perceptrons.} \citep{vogn, shridhar2019comprehensive}.
FROMP performs consistently better across all tasks (except the first task where it is close to the best).
It also improves over the lower limit (`separate tasks') by a large margin.
In fact, on tasks 4-6, FROMP matches the performance to the network trained jointly on all tasks, which implies that there it completely avoids forgetting. 
The average performance over all tasks is also the best (see the `Avg' column).

\cref{fig:mem_vs_acc_splitcifar} shows the performance with respect to the number of memorable past examples. 
Similarly to \cref{fig:permuted_mem_vs_acc}, carefully selecting memorable example improves the performance, especially when the number of memorable examples is small.
For example, with 10 such memorable examples, a careful selection in FROMP increases the average accuracy to $70\%$ from $45\%$ obtained by FRORP.
Including the kernel in FROMP here unfortunately does not improve significantly over FROMP-$L_2$, unlike the MNIST experiment.
\cref{fig:memorable_past_cifar} shows a few images with most and least memorable past examples where we again see that the most memorable might be more difficult to classify.

Finally, we analyse the forward and backward transfer obtained by FROMP. Forward transfer means the accuracy on the \emph{current} tasks increases as number of past tasks increases, while backward transfer means the accuracy on the \emph{previous} tasks increases as more tasks are observed. As discussed in \cref{app:metrics}, we find that, for Split CIFAR, FROMP's forward transfer is much better than VCL and EWC, while its backward transfer is comparable to EWC. 
We define a forward transfer metric as the average improvement in accuracy on a new task over a model trained \emph{only} on that task (see \cref{app:metrics} for an expression). A higher value is better and quantifies the performance gain by observing past tasks.
FROMP achieves $6.1\pm0.7\%$, a much higher value compared to $0.17\pm0.9\%$ obtained with EWC and $1.8\pm3.1\%$ with VCL+coresets.
For backward transfer, we used the BWT metric defined in \citet{lopez2017gradient} which roughly captures the difference in accuracy obtained when a task is first trained and its accuracy after the final task. Again, higher is better and quantifies the gain obtained with the future tasks.
Here, FROMP has a score of $-2.6\pm0.9\%$, which is comparable to EWC's score of $-2.3\pm1.4\%$ but better than VCL+coresets which obtains $-9.2\pm1.8\%$. 
Performance of FROMP is summarised in Table \ref{tab:fwd_bwd_metrics}.

\begin{table*}[th]
    \centering
    \begin{tabular}{cccccc}
        \toprule 
        \multicolumn{2}{c}{\textbf{Split MNIST}} & \multicolumn{2}{c}{\textbf{Permuted MNIST}} & \multicolumn{2}{c}{\textbf{Split CIFAR}}\\
        FWD & BWD & FWD & BWD & FWD & BWD\\
        \hline
            $-0.07\pm0.05\%$    & $-0.5\pm0.2\%$    & $-1.9\pm0.1\%$    & $-1.0\pm0.1\%$    & $6.1\pm0.7\%$    & $-2.6\pm0.9\%$ \\
        \bottomrule
    \end{tabular}
        \caption{Forward and Backward transfer metrics (see \cref{app:metrics} for precise definitions and more results) for FROMP on benchmarks. Higher is better.
    }
    \label{tab:fwd_bwd_metrics}
\end{table*}

%% file: sections/5discussion.tex
We propose FROMP, a functional-regularisation approach for continual learning while avoiding catastrophic forgetting. FROMP uses a Gaussian Process formulation of neural networks to convert weight-space distributions into function-space.
With this formulation, we proposed ways to identify relevant \emph{memorable past} examples, and functionally regularise the training of neural network weights. FROMP achieves state-of-the-art performance across benchmarks.

This paper takes the first step in a direction to combine the ideas from neural network and GP communities, while maintaining the simplicity of the training methods. There are plenty of future directions to explore. Would using VI instead of a Laplace approximation result in better accuracy? What are some ways to choose memorable examples? Is there a common principle behind them? How many memorable examples should one use, and how can we ensure that increasing their numbers substantially increases the performance? Do we obtain improvements when we relax some assumptions, and what kind of improvements? Will this approach work at large scale, e.g., on the ImageNet dataset? Are there better methods to automatically detect task boundaries? And finally how can all of these ideas fit in a Bayesian framework, and can we obtain theoretical guarantees for such methodologies?

These are some of the questions of future interest.
We believe that functional regularisation is ultimately how we want to train deep learning algorithms. We hope that the methods discussed in this paper open new methodologies for knowledge transfer in deep learning. 

%% file: sections/appendix/app_A.tex
\section{Deep Networks to Functional Priors with DNN2GP} \label{app:binary_derivation}

\subsection{GP posteriors from the Minimiser of Linear Model} \label{app:GP_predictive}
The posterior distribution of a linear model induces a GP posterior as shown by \citet{rasmussen2006gaussian}. We discuss this in detail now for the following linear model discussed in \cref{sec:dnn2gp}:
\begin{align}
y_i = f_\param(\vx_i) + \epsilon_i, \; \textrm{ where } f_\param(\vx_i) := \vphi(\vx_i)^\top\vparam, \; \epsilon_i \sim \gauss(\epsilon_i| 0, \Lambda^{-1}), \; \textrm{ and } \vparam \sim \gauss(\vparam|0, \delta^{-1}\vI_P)
\end{align}
with a feature map $\vphi(\vx)$. \citet{rasmussen2006gaussian} show that the predictive distribution for a test input $\vx$ takes the following form (see Equation 2.11 in their book):
\begin{align}
    p(f(\vx)|\vx, \data) = \gauss( f(\vx) \, | &\, \Lambda \vphi(\vx)^\top \vA^{-1} \vPhi \vy,\,\,\, \vphi(\vx)^\top \vA^{-1} \vphi(\vx)), \nonumber\\
    &\textrm{ where } \vA := \sum_i \vphi(\vx_i)\, \Lambda\, \vphi(\vx_i)^\top + \delta \vI_P.
    \label{eq:predGP_lin}
\end{align}
where $\data$ is set of training points $\{y_i,\vx_i\}$ for $i$, and $\vPhi$ is a matrix with $\vphi(\vx_i)$ as columns.

\citet{rasmussen2006gaussian} derive the above predictive distribution by using the weight-space posterior $\gauss(\vparam|\vparam_\textrm{lin}, \vSigma_{\textrm{lin}})$ with the mean and covariance defined as below:
\begin{align}
    \vparam_\textrm{lin} := \Lambda \vA^{-1} \vPhi \vy, \quad 
    \vSigma_\textrm{lin} := \vA^{-1}.
    \label{eq:post_lin}
\end{align}
The mean $\vparam_{lin}$ is also the minimiser of the least-squares loss and $\vA$ is the hessian at that solution. 

\citet{rasmussen2006gaussian} show that the predictive distribution in \cref{eq:predGP_lin} corresponds to a GP posterior with the following mean and covariance functions:
\begin{align}
    m_\textrm{lin}(\vx) &= \Lambda \vphi(\vx)^\top \vA^{-1} \vPhi \vy = \vphi(\vx)^\top \vparam_\textrm{lin} = f_{\param_\textrm{lin}}(\vx) \label{eq:lin_gp_mean},\\ 
    \kappa_\textrm{lin}(\vx,\vx') &:= \vphi(\vx)^\top\, \vSigma_\textrm{lin}\, \vphi(\vx'), \label{eq:lin_gp_cov}
\end{align}
This is the result shown in \cref{eq:lin_gp} in \cref{sec:dnn2gp}. We can also write the predictive distribution of the observation $y = f(\vx) + \epsilon$ where $\epsilon \sim\gauss(0,\Lambda^{-1})$ as follows:
\begin{align}
    p(y|\vx, \data) = \gauss( y \, | &\, \underbrace{f_{\param_\textrm{lin}}(\vx)}_{m_\textrm{lin}(\mathbf{x})},\,\,\, \underbrace{\vphi(\vx)^\top \vSigma_\textrm{lin} \vphi(\vx)}_{\kappa_\textrm{lin}(\mathbf{x},\mathbf{x})} + \Lambda^{-1}), \nonumber\\
    &\textrm{ where } \vSigma_\textrm{lin}^{-1} := \sum_i \vphi(\vx_i)\, \Lambda\, \vphi(\vx_i)^\top + \delta \vI_P. 
    \label{eq:predGP_lin_y}
\end{align}
We will make use of \cref{eq:lin_gp_mean,eq:lin_gp_cov,eq:predGP_lin_y} to write the mean and covariance function of the posterior approximation for neural networks, as shown in the next section.

\subsection{GP Posteriors from the Minimiser of Neural Networks}
\label{app:ste}
\citet{nn2gp} derive GP predictive distributions for the minimisers of a variety of loss functions in Appendix B of their paper. We describe these below along with the resulting GP posteriors. Throughout, we denote a minimiser of the loss by $\vparam_*$.

{\bf A regression loss:} For a regression loss function $\loss(y, f) := \half \Lambda(y-f)^2$, they derive the following expression for the predictive distribution for the observations $y$ (see Equation 44, Appendix B.2 in their paper):
\begin{align}
    \hat{p}(y|\vx, \data) := \gauss( y \, | &\, f_{\param_*}(\vx),\,\,\, \vJ_{\param_*}(\vx) \vSigma_* \vJ_{\param_*}(\vx)^\top + \Lambda^{-1}), \nonumber\\
    &\textrm{ where } \vSigma_*^{-1} := \sum_i \vJ_{\param_*}(\vx_i)^\top \, \Lambda\, \vJ_{\param_*}(\vx_i) + \delta \vI_P.
    \label{eq:predGP_dnn2gp}
\end{align}
We use $\hat{p}(y|\vx,\data)$ since this predictive distribution is not exact and is obtained using a type of Laplace approximation.
Comparing this to \cref{eq:predGP_lin_y}, we can write the mean and covariance functions in a similar fashion as \cref{eq:lin_gp_mean,eq:lin_gp_cov}:
\begin{align}
    m_{\param_*}(\vx) &:= f_{\param_*}(\vx), \quad
    \kappa_{\param_*}(\vx,\vx') := \vJ_{\param_*}(\vx)\, \vSigma_*\, \vJ_{\param_*}(\vx')^\top .
\end{align}
This is the result shown in \cref{eq:laplace_kernel_post_gp} in \cref{sec:dnn2gp}.

{\bf A binary classification loss:} A similar expression is available for binary classification with $y\in\{0,1\}$, considering the loss $\loss(y,f) := -y\log\sigma(f) - (1-y)\log(1-\sigma(f)) = -yf + \log (1+e^f)$ where $\sigma(f) := 1/(1+e^{-f})$ is the sigmoid function. See Equation 48, Appendix B.2 in \citet{nn2gp}. The predictive distribution is given as follows:
\begin{align}
    \hat{p}(y|\vx, \data) := \gauss( y \, | &\, \sigma(f_{\param_*}(\vx)),\,\,\, \Lambda_{\param_*}(\vx)\, \vJ_{\param_*}(\vx)\, \vSigma_*\, \vJ_{\param_*}(\vx)^\top \Lambda_{\param_*}(\vx) + \Lambda_{\param_*}(\vx)), \nonumber\\
    &\textrm{ where } \vSigma_*^{-1} := \sum_i \vJ_{\param_*}(\vx_i)^\top \, \Lambda_{\param_*}(\vx_i)\, \vJ_{\param_*}(\vx_i) + \delta \vI_P.
    \label{eq:predGP_dnn2gp_bin}
\end{align}
where $\vLambda_{\param_*}(\vx) := \sigma\rnd{f_{\param_*}(\vx)} \sqr{1- \sigma\rnd{f_{\param_*}(\vx)}}$. 
The predictive distribution does not respect the fact that $y$ is binary and treats it like a Gaussian. This makes it comparable to \cref{eq:predGP_lin_y}. Comparing the two, we can conclude that the above corresponds to the predictive posterior distribution of a GP regression model with $y = f(\vx) + \epsilon$ where $\epsilon \sim \gauss(0, \Lambda_{\param_*}(\vx))$ with the mean and covariance function as shown below:
\begin{align}
    m_{\param_*}(\vx) &:= \sigma(f_{\param_*}(\vx)), \quad
    \kappa_{\param_*}(\vx,\vx') := \Lambda_{\param_*}(\vx)\, \vJ_{\param_*}(\vx)\, \vSigma_*\, \vJ_{\param_*}(\vx')^\top \Lambda_{\param_*}(\vx) .
\end{align}
This is the result used in \cref{eq:mK_bin} in \cref{sec:final_algo} for binary classification. A difference here is that the mean function is passed through the sigmoid function and the covariance function has $\Lambda_{\param_*}(\vx)$ multiplied on the both sides. These changes appear because of the nonlinearity in the loss function introduced due to the sigmoid link function.

{\bf A multiclass classification loss:} The above result straightforwardly extends to the multiclass classification case by using multinomial-logit likelihood (or softmax function). For this the loss can be written as follows:
\begin{align}
    \loss(\vy,\vf) = -\vy^\top \mysoftmax(\vf) + \log \rnd{1+ \sum_{k=1}^{K-1} e^{f_k}}, \textrm{ where $k$'th element of $\mysoftmax(\vf)$ is } \frac{e^{f_j}}{1+\sum_{k=1}^{K-1} e^{f_k}} 
\end{align}
where the number of categories is equal to $K$, $\vy$ is a one-hot-encoding vector of size $K-1$, $\vf$ is $K-1$ length output of the neural network, and $\mysoftmax(\vf)$ is the softmax operation which maps a $K-1$ length real vector to a $K-1$ dimensional vector with entries in the open interval $(0, 1)$. The encoding in $K-1$ length vectors ignores the last category which then ensures identifiability \citep{train2009discrete}.
In a similar fashion to the binary case, the predictive distribution of the $K-1$ length output $\vy$ for an input $\vx$ can be written as follows:
\begin{align}
    \hat{p}(\vy|\vx, \data) := \gauss( \vy \, | &\, \mysoftmax(\vf_{\param_*}(\vx)),\,\,\, \vLambda_{\param_*}(\vx)\, \vJ_{\param_*}(\vx)\, \vSigma_*\, \vJ_{\param_*}(\vx)^\top \vLambda_{\param_*}(\vx)^\top + \vLambda_{\param_*}(\vx)), \nonumber\\
    &\textrm{ where } \vSigma_*^{-1} := \sum_i \vJ_{\param_*}(\vx_i)^\top \, \vLambda_{\param_*}(\vx_i)\, \vJ_{\param_*}(\vx_i) + \delta \vI_P.
    \label{eq:predGP_dnn2gp_multi}
\end{align}
where $\vLambda_{\param_*}(\vx) := \mysoftmax\rnd{\vf_{\param_*}(\vx)} \sqr{1- \mysoftmax\rnd{\vf_{\param_*}(\vx)}}^\top$ is a $(K-1)\times (K-1)$ matrix and $\vJ_{\param_*}(\vx)$ is the $(K-1)\times P$ Jacobian matrix. 
The mean function in this case is a $K-1$ length matrix and the covariance function is a square matrix of size $K-1$. Their expressions are shown below:
\begin{align}
    \vm_{\param_*}(\vx) &:= \mysoftmax(\vf_{\param_*}(\vx)), \quad
    \boldsymbol{K}_{\param_*}(\vx,\vx') := \vLambda_{\param_*}(\vx)\, \vJ_{\param_*}(\vx)\, \vSigma_*\, \vJ_{\param_*}(\vx')^\top \vLambda_{\param_*}(\vx') .
\end{align}

{\bf General case: } The results above hold for a generic loss function derived from a generalised linear model (GLM) with an invertible function $\vh(\vf)$, e.g., $\loss(\vy,\vf) := - \log p(\vy|\vh(\vf))$. For example, for a Bernoulli distribution, the link function $h(f)$ is equal to $\sigmoid$. In the GLM literature, $\vh^{-1}$ is known as the link function. Given such a loss, the only quantity that changes in the above calculations is $\vLambda_{\param_*}(\vx,\vy) := \nabla_{ff}^2 \loss(\vy,\vf)$, which is the second derivative of the loss with respect to $\vf$, and might depend both on $\vx$ and $\vy$.

\subsection{GP Posterior from the Iterations of a Neural-Network Optimiser} \label{app:gp_predictive_von}
The results of the previous section hold only at a minimiser $\vparam_*$. \citet{nn2gp} generalise this to iterations of optimisers. They did this for a  variational inference algorithm and also for its deterministic version that resembles RMSprop. We now describe these two versions. We will only consider binary classification using the setup described in the previous section. The results can be easily generalised to multiclass classification.

{\bf GP posterior from iterations of a variational inference algorithm: }
Given a Gaussian variational approximation $q_j(\vparam) := \gauss(\vparam|\vmu_j, \vSigma_j)$ at iteration $j$, \citet{nn2gp} used a natural-gradient variational inference algorithm called the variational-online Newton (VON) method~\cite{vadam}.
Given a $q_j(\vparam)$, the algorithm proceeds by first sampling $\vparam_j \sim q_j(\vparam)$, and then updating the variational distribution.
Surprisingly, the procedure used to derive a GP predictive distribution for the minimiser generalises to this update too. 
An expression for the predictive distribution is given below:
\begin{align}
    \hat{p}_{j+1}(y|\vx, \data) &:= \gauss( y \, | \, \sigma(f_{\param_j}(\vx)),\,\,\, \Lambda_{\param_{j}}(\vx)\, \vJ_{\param_{j}}(\vx)\, \vSigma_{j}\, \vJ_{\param_{j}}(\vx)^\top \Lambda_{\param_{j}}(\vx) + \Lambda_{\param_{j}}(\vx)^{-1}), \label{eq:pred_GP_var} \\
    \textrm{ where } \vSigma_{j+1}^{-1} &:= (1-\beta_j) \vSigma_{j}^{-1} + \beta_j \sqr{ \sum_i \vJ_{\param_{j}}(\vx_i)^\top \, \Lambda_{\param_{j}}(\vx_i)\, \vJ_{\param_{j}}(\vx_i) + \delta \vI_P},\\
    \vmu_{j+1} &:= \vmu_{j} - \beta_j \vSigma_{j+1} \sqr{N \nabla_\param \bar{\loss}(\vparam_j) + \delta\vmu_j},
    \label{eq:predGP_dnn2gp_bin_iter}
\end{align}
where $\bar{\loss}(\vparam) := \frac{1}{N} \sum_{i=1}^N \loss(y_i, f_\param(\vx_i))$.
The predictive distribution takes the same form as before, but now the covariance and mean are updated according to the VON updates. The VON updates are essential to ensure the validity of the GP posterior, however, as \citet{nn2gp} discuss, the RMSprop/Adam have similar update which enable us to apply the above results even when running such algorithms. We describe this next.  

{\bf GP posterior from iterations of RMSprop/Adam: } \citet{nn2gp} propose a deterministic version of the above update where $\vparam_j$ is not sampled from $q_j(\vparam)$ rather is set to be equal to $\vmu_j$, i.e., $\vparam_j = \vmu_j$. This gives rise to the following update:
\begin{align}
    \vSigma_{j+1}^{-1} &\leftarrow (1-\beta_j) \vSigma_{j}^{-1} + \beta_j \sqr{ \sum_i \vJ_{\param_{j}}(\vx_i)^\top \, \Lambda_{\param_{j}}(\vx_i)\, \vJ_{\param_{j}}(\vx_i) + \delta \vI_P}, \label{eq:rmsprop_like_1}\\
    \vparam_{j+1} &\leftarrow \vparam_{j} - \beta_j \vSigma_{j+1} \sqr{ N \nabla_\param \bar{\loss}(\vparam_j) + \delta\vparam_j }\label{eq:rmsprop_like_2},
\end{align}
with the variational approximation defined as $q_j(\vparam) := \gauss(\vparam|\vparam_j, \vSigma_j)$. The form of the predictive distribution remains the same as \cref{eq:pred_GP_var}.

As discussed in \citet{vadam}, the above algorithm can be made similar to RMSprop by using a diagonal covariance.
By reparameterising the diagonal of $\vSigma^{-1}$ as $\vs + \delta\vone$ where $\vs$ is an unknown vector, we can rewrite the updates to update $\vmu$ and $\vs$. This can then be written in a form similar to RMSprop as shown below: 
\begin{align}
    \vs_{j+1} &\leftarrow (1-\beta_j) \vs_{j} + \beta_j \Big[ \sum_{i} \Lambda_{\param_j}(\vx_i) \sqr{ \vJ_{\param_j}(\vx_i) \circ \vJ_{\param_j}(\vx_i)}^\top \Big] \label{eq:s_vec}\\
    \vparam_{j+1} &\leftarrow \vparam_{j} - \beta_t \frac{1}{\vs_{j+1} + \delta\vone} \circ \sqr{ N\nabla_\param \bar{\loss}(\vparam_t) + \delta\vparam_j}, \label{eq:rmsprop_like_mu}
\end{align}
where $\circ$ defines element-wise product of two vectors, and the diagonal of $\vSigma_{j+1}^{-1}$ is equal to $(\vs_{j+1} + \delta\vone)$. This algorithm differs from RMSprop in two ways. First, the scale vector $\vs_j$ is updated using the sum of the square of the Jacobians instead of the square of the mini-batch gradients. Second, there is no square-root in the preconditioner for the gradient in the second line. This algorithm is the diagonal version of the Online Generalised Gauss-Newton (OGGN) algorithm discussed in \citet{nn2gp}.  

In practice, we ignore these two differences and employ the RMSprop/Adam update instead. As a consequence the variance estimates might not be very good during the iteration, even though the fixed-point of the algorithm is not changed \citep{vadam}. This is the price we pay for the convenience of using RMSprop/Adam. We correct the approximation after convergence of the algorithm by recomputing the diagonal of the covariance according to \cref{eq:s_vec}. Denoting the converged solution by $\vparam_*$, we compute the diagonal $\vv_*$ of the covariance $\vSigma_*$ as shown below:
\begin{align}
    \vv_* = \vone / \Big[ \delta\vone + \sum_{i=1}^N \Lambda_{\param_*}(\vx_i) \sqr{ \vJ_{\param_*}(\vx_i) \circ \vJ_{\param_*}(\vx_i)}^\top \Big], 
    \label{eq:cov_bin_app}
\end{align}

\section{Detailed Derivation of FROMP Algorithm} \label{app:derivation}

In this section, we provide further details on \cref{sec:putting together}.

\begin{align}
    \mathcal{L}(q(\vparam)) := \myexpect_{q(\param)} \left[ \frac{N}{\tau} \bar{\loss}_t(\vparam) + \log q(\vparam) \right] - \, & \textcolor{red}{\myexpect_{\tilde{q}_{\param_t}(\rf)} \sqr{\log \tilde{q}_{\param_{t-1}}(\vf)}} , \nonumber\\
    &\textrm{ where } \vparam_t \sim q(\vparam) \textrm{ and } \vparam_{t-1} \sim q_{t-1}(\vparam). \label{eq:fromp_loss_app}
\end{align}
Optimising this objective requires us to obtain the GP posterior $\tilde{q}_{\param_t}(\vf)$. This can be easily done applying the DNN2GP result from \cref{eq:pred_GP_var} to this loss function.
The VON update for the objective above takes the following form:
\begin{align}
    \vSigma^{-1} &\leftarrow (1-\beta) \vSigma^{-1} + \beta \sqr{ \sum_i \vJ_{\param_{t}}(\vx_i)^\top \, \Lambda_{\param_t}(\vx_i)\, \vJ_{\param_{t}}(\vx_i) - 2\nabla_{\Sigma} \textcolor{red}{\myexpect_{\tilde{q}_{\param_t}(\rf)} \sqr{\log \tilde{q}_{\param_{t-1}}(\vf)}} }, \label{eq:fromp_von_sigma_app}\\
    \vmu &\leftarrow \vmu - \beta \vSigma \sqr{\frac{N}{\tau} \nabla_\param \bar{\loss}_t(\vparam_t) - \nabla_{\mu} \textcolor{red}{\myexpect_{\tilde{q}_{\param_t}(\rf)} \sqr{\log \tilde{q}_{\param_{t-1}}(\vf)}}} .
    \label{eq:fromp_von_mu_app}
\end{align}
where $\bar{\loss}_t(\vparam) := \frac{1}{N} \sum_{i\in\data_t} \loss(y_i, f_\param(\vx_i))$ and we have ignored the iteration subscript to simplify notation.

Using the $\vmu$ and $\vSigma$ obtained with this iteration, we can define the following GP predictive posterior at a sample $\vparam_t\sim q(\vparam)$:
\begin{align}
    \hat{p}_t(y|\vx, \data) := \gauss( y \, | \, \sigma(f_{\param_{t}}(\vx)),\,\,\, &\Lambda_{\param_{t}}(\vx)\, \vJ_{\param_{t}}(\vx)\, \vSigma\, \vJ_{\param_{t}}(\vx)^\top \Lambda_{\param_{t}}(\vx) + \Lambda_{\param_{t}}(\vx)^{-1}), 
\end{align}
Comparing this to \cref{eq:pred_GP_var}, we can write the mean and covariance function as follows:
\begin{align}
    m_{\param_t}(\vx) &:= \sigma(f_{\param_t}(\vx)), \quad
    \kappa_{\param_t}(\vx,\vx') := \Lambda_{\param_t}(\vx)\, \vJ_{\param_t}(\vx)\, \vSigma\, \vJ_{\param_t}(\vx')^\top \Lambda_{\param_t}(\vx) .
\end{align}
The mean vector obtained by concatenating $m_{\param_t}(\vx)$ at all $\vx\in\memory$ is denoted by $\vm_t$. Similarly, the covariance matrix $\vK_t$ is defined as the matrix with $ij$'th entry as $\kappa_{\param_t}(\vx_i,\vx_j)$. The corresponding mean and covariance obtained from samples from $q_{t-1}(\vparam)$ are denoted by $\vm_{t-1}$ and $\vK_{t-1}$.

Given these quantities, the functional regularisation term has an analytical expression given as follows:
\begin{align}
    \myexpect_{\tilde{q}_{\param_t}(\rf)} \sqr{\log \tilde{q}_{\param_{t-1}}(\vf)} &=-\half \sqr{\trace{(\vK_{t-1}^{-1} \vK_{t})} + (\vm_t - \vm_{t-1})^\top \vK_{t-1}^{-1} (\vm_t - \vm_{t-1})},
    \label{eq:func_reg_app}
\end{align}
correct to a constant.
Our goal is to obtain the derivative of this term with respect to $\vmu$ and $\vSigma$. Both $\vm_t$ and $\vK_t$ are functions of $\vmu$ and $\vSigma$ through the sample $\vparam_t = \vmu + \vSigma^{1/2} \vepsilon$ where $\vepsilon \sim \gauss(0,\vI)$. Therefore, we can compute these derivative using the chain rule.

We note that the resulting algorithm is costly for large problems, and propose five approximations to reduce the computation cost, as described below.

\textbf{Approximation 1:} Instead of sampling $\vparam_{t-1}$, we set $\vparam_{t-1} = \vmu_{t-1}$ which is the mean of the posterior approximation $q_{t-1}(\vparam)$ until task $t-1$. Therefore, we replace $\myexpect_{\tilde{q}_{\param_t}(\rf)} \sqr{\log \tilde{q}_{\param_{t-1}}(\vf)}$ by $\myexpect_{\tilde{q}_{\param_t}(\rf)} \sqr{\log \tilde{q}_{{\color{red} \mu_{t-1}}}(\vf)}$. This affects the mean $\vm_{t-1}$ and $\vK_{t-1}$ in \cref{eq:func_reg_app}.

\textbf{Approximation 2:} When computing the derivation of the functional regulariser, we will ignore the derivative with respect to $\vK_t$ and only consider $\vm_t$. Therefore, the derivatives needed for the update in \cref{eq:fromp_von_sigma_app,eq:fromp_von_mu_app} can be approximated as follows:
\begin{align}
    \nabla_{\mu} \myexpect_{\tilde{q}_{\param_t}(\rf)} \sqr{\log \tilde{q}_{\param_{t-1}}(\vf)} &\approx -\sqr{\nabla_{\mu} \vm_t} \vK_{t-1}^{-1} (\vm_t - \vm_{t-1}), \label{eq:grad_mu}\\
    \nabla_{\Sigma} \myexpect_{\tilde{q}_{\param_t}(\rf)} \sqr{\log \tilde{q}_{\param_{t-1}}(\vf)} &\approx -\sqr{\nabla_{\Sigma} \vm_t} \vK_{t-1}^{-1} (\vm_t - \vm_{t-1}). \label{eq:grad_Sigma}
\end{align}
This avoids having to calculate complex derivatives (e.g., derivatives of Jacobians).

\textbf{Approximation 3:} Instead of using the full $\vK_{t-1}$, we factorise it across tasks, i.e., we approximate it by a block-diagonal matrix containing the kernel matrix $\vK_{t-1,s}$ for all past tasks $s$ as the diagonal.
This makes the cost of inversion linear in the number of tasks.

\textbf{Approximation 4:} 
Similarly to \cref{eq:rmsprop_like_1,eq:rmsprop_like_2}, we use a deterministic version of the VON update by setting $\vparam_t = \vmu$, which corresponds to setting the random noise $\epsilon$ to zero in $\vparam_t = \vmu + \vSigma^{1/2}\vepsilon$.
This approximation simplifies the gradient computation in \cref{eq:grad_mu,eq:grad_Sigma}, since now the gradient with respect to $\vSigma$ is zero. For example, in the binary classification case, $m_\mu(\vx) := \sigma(f_\mu(\vx))$, which does not depend on $\vSigma$. The gradient of $\vm_t$ with respect to $\vmu$ is given as follows using the chain rule (here $\vm_{t,s}$ is the sub-vector of $\vm_t$ corresponding to the task $s$). 
\begin{align}
    \nabla_{\mu}\vm_{t,s}[i] = \nabla_{\mu} \sqr{  \sigma\rnd{f_{\mu}(\vx_i)}}=  \Lambda_{\mu}(\vx_i)\, \vJ_{\mu}(\vx_i)^\top, \textrm{ where } \vx_i \in \memory_s, \label{eq:grad_m_bin_app}
\end{align}
and where the second equality holds for canonical link functions. With these simplifications, we can write the VON update as follows:
\begin{align}
    \vSigma^{-1} &\leftarrow (1-\beta) \vSigma^{-1} + \beta \sqr{ \sum_i \vJ_{\mu}(\vx_i)^\top \, \Lambda_{\mu}(\vx_i)\, \vJ_{\mu}(\vx_i)},\\
    \vmu &\leftarrow \vmu - \beta \vSigma \sqr{\frac{N}{\tau} \nabla_\mu \bar{\loss}_t(\vmu) + \sum_{s=1}^{t-1} \sqr{\nabla_{\mu} \vm_{t,s}} \vK_{t-1, s}^{-1} (\vm_{t,s} - \vm_{t-1, s})} .
\end{align}

    {\bf Approximation 5:} Similarly to \cref{eq:s_vec,eq:rmsprop_like_mu}, our final approximation is to use a diagonal covariance $\vSigma$ and replace the above update by an RMSprop-like update where we denote $\vmu$ by $\vparam$:
\begin{align}
    \vs &\leftarrow (1-\beta) \vs + \beta \sqr{ \sum_{i} \Lambda_{\param}(\vx_i) \sqr{ \vJ_{\param}(\vx_i) \circ \vJ_{\param}(\vx_i)}^\top}, \label{eq:rmsprop_like_final_1}\\
    \vparam &\leftarrow \vparam - \beta \frac{1}{\vs + \delta\vone} \circ  \sqr{\frac{N}{\tau} \nabla_\param \bar{\loss}_t(\vparam) + \sum_{s=1}^{t-1} \sqr{\nabla_{\param} \vm_{t,s}} \vK_{t-1, s}^{-1} (\vm_{t,s} - \vm_{t-1, s})} , \label{eq:rmsprop_like_final_2}
\end{align}
where we have added a regulariser $\delta$ to $\vs$ in the second line to avoid dividing by zero. Previously \citep{vadam}, this regulariser was the prior precision. Ideally, when using a functional prior, we would replace this by another term. However, this term was ignored by making Approximation~4, and we use $\delta$ instead.
The final Gaussian approximation is obtained with the mean equal to $\vparam$ and covariance is equal to a diagonal matrix with $1/(\vs + \delta\vone)$ as its diagonal.

It is easy to see that the solutions found by this algorithm is the fixed point of this objective:
\begin{align}
    \label{eq:fromp_final_reg_app}
   \min_{\param} N\bar{\loss}_t(\vparam) + \half \tau \sum_{s=1}^{t-1} (\vm_{t,s} - \vm_{t-1,s})^\top \vK_{{t-1},s}^{-1} (\vm_{t,s} - \vm_{t-1,s}) ,
\end{align}
Ultimately, this is an approximation of the objective given in \cref{eq:fromp_loss_app}, and is computationally cheaper to optimise.

We follow the recommendations of \citet{nn2gp} and use RMSprop/Adam instead of \cref{eq:rmsprop_like_1,eq:rmsprop_like_2}. This algorithm still optimises the objective given in \cref{eq:fromp_final_reg_app}, but the estimate of the covariance is not accurate. We correct the approximation after convergence of the algorithm by recomputing the diagonal of the covariance according to \cref{eq:rmsprop_like_final_1}. Denoting the converged solution by $\vparam_*$, we compute the diagonal $\vv_*$ of the covariance $\vSigma_*$ as shown below:
\begin{align}
    \vv_* = \vone / \Big[ \delta\vone + \sum_{i=1}^N \Lambda_{\param_*}(\vx_i) \sqr{ \vJ_{\param_*}(\vx_i) \circ \vJ_{\param_*}(\vx_i)}^\top \Big], 
    \label{eq:cov_bin_fromp}
\end{align}

%% file: sections/appendix/derivation_multiclass.tex
\section{Multiclass setting} \label{app:multiclass}

When there are more than two classes per task, we need to use multiclass versions of the equations presented so far. We still make the same approximations as described in \cref{app:derivation}.

\textbf{Reducing Complexity in the Multiclass setting:} We could use the full multiclass version of the GP predictive (\cref{eq:predGP_dnn2gp_multi}), but this is expensive. To keep computational complexity low, we employ an individual GP over each of the $K$ classes seen in a previous task, and treat the GPs as independent. 

We have $K$ separate GPs. Let $\ry^{(k)}$ be the $k$-th item of $\vy$. Then the predictive distribution over each $\ry^{(k)}$ for an input $\vx$ is:
\begin{align}
    \hat{p}(\ry^{(k)}|\vx, \data) := \gauss \big( \ry^{(k)} \, | \, \mysoftmax(\vf_{\param_*}(\vx))^{(k)},\,\,\, \vLambda_{\param_*}(\vx)^{(k)}\, \vJ_{\param_*}(\vx)\, \vSigma_*\, \vJ_{\param_*}(\vx)^\top &\vLambda_{\param_*}(\vx)^{(k)\top} \nonumber\\
    &+ \Lambda_{\param_*}(\vx)^{(k,k)} \big), 
    \label{eq:predGP_dnn2gp_multi_approx}
\end{align}
where $\mysoftmax(\vf_{\param_*}(\vx))^{(k)}$ is the k-th output of the softmax function, $\vLambda_{\param_*}(\vx)^{(k)}$ is the $k$-th row of the Hessian matrix and $\Lambda_{\param_*}(\vx)^{(k,k)}$ is the $k,k$-th element of the Hessian matrix. The Jacobians $\vJ_{\param_*}(\vx)$ are now of size $K\times P$. Note that we have allowed $\mysoftmax$ and $\Lambda_{\param_*}(\vx)$ to be of size $K$ instead of $K-1$. This is because we are treating the $K$ GPs separately.

The kernel matrix $\vK_{t-1}$ is now a block diagonal matrix for each previous task's classes. This allows us to only compute inverses of each block diagonal (size $M\times M$), repeated for each class in each past task ($K(t-1)$ times), where $M$ is the number of memorable past examples in each task. This changes computational complexity to be linear in the number of classes per task, $K$, compared to \cref{sec:final_algo} (which has analysis for binary classification for each task).

When choosing a memorable past (the subset of points to regularise function values over) for the logistic regression case, we can simply sort the $\Lambda_{\param_*}(\vx_i)$'s for all $\{\vx_i\} \in \data_t$ and pick the largest, as explained in \cref{sec:choosing mem}. In the multiclass case, these are now $K\times K$ matrices $\vLambda_{\param_*}(\vx_i)$. We instead sort by $\trace(\vLambda_{\param_*}(\vx_i))$ to select the memorable past examples.

\textbf{FROMP for multiclass classification:} 
The solutions found by the multiclass algorithm is the fixed point of this objective (compare with \cref{eq:fromp_final_reg_app}):
\begin{align}
    \label{eq:fromp_multi_final_reg_app}
   \min_{\param} N\bar{\loss}_t(\vparam) + \half \tau  \sum_{s=1}^{t-1} \sum_{k\in C_s} (\vm_{t,s,k} - \vm_{t-1,s,k})^\top \vK_{{t-1},s,k}^{-1} (\vm_{t,s,k} - \vm_{t-1,s,k}) ,
\end{align}
where we define $C_s$ as the set of classes $k$ seen in previous task $s$, $\vm_{t,s,k}$ is the vector of $m_{\param_t}(\vx)$ for class $k$ evaluated at the memorable points $\{\vx_i\} \in \memory_s$, $\vm_{t-1,s,k}$ is the vector of $m_{\param_{t-1}}(\vx)$ for class $k$, and $\vK_{t-1,s,k}$ is the kernel matrix from the previous task just for class $k$, always evaluated over just the memorable points from previous task $s$. By decomposing the last term over individual outputs and over the memorable past from each task, we have reduced the computational complexity per update.

%% file: sections/appendix/functional_approx.tex
\section{Functional prior approximation}\label{app:functional_approx}

We discuss why replacing weight space integral by a function space integral, as done below, results in an approximation: 
\begin{align}
    \myexpect_{q(\param)} [\log q_{t-1}(\vparam)] &\approx \myexpect_{\tilde{q}_{\param_t}(\rf)} \sqr{\log \tilde{q}_{\param_{t-1}}(\vf)} ,  \nonumber 
\end{align}
A change of variable in many cases results in an equality, e.g., for $\vf = \vX\vparam$ with a matrix $\vX$ and given any function $h(\vf)$, we can express the weight space integral as the function space integral: 
\begin{align}
    \int h(\vX\vparam) \gauss(\vparam|\vmu,\vSigma) d\vparam = \int h(\vf) \gauss(\vf|\vX\vmu, \vX\vSigma\vX^\top) d\vf.
\end{align}
Unfortunately, $\log q_{t-1}(\vparam)$ can not always be written as a function of $\vf := \vJ_{\param_t} \vparam$. Therefore, the change of variable does not result in an equality. For our purpose, as long as the approximations provide a reasonable surrogate for optimisation, the approximation is not expected to cause issues. 

%% file: sections/appendix/metrics_app.tex
\section{Further details on continual learning metrics reported}
\label{app:metrics}

We report a backward transfer metric and a forward transfer metric on Split CIFAR (higher is better for both). The backward transfer metric is exactly as defined in \citet{lopez2017gradient}. The forward transfer metric is a measure of how well the method uses previously seen knowledge to improve classification accuracy on newly seen tasks. Let there be a total of $T$ tasks. Let $R_{i,j}$ be the classification accuracy of the model on task $t_j$ after training on task $t_i$. Let $R^\textrm{ind}_i$ be the classification accuracy of an independent model trained only on task $i$. Then,

\begin{align*}
    \textrm{Backward Transfer, BWT} = \frac{1}{T-1} \sum_{i=1}^{T-1} R_{T,i} - R_{i,i}, \\
    \textrm{Forward Transfer, FWT} = \frac{1}{T-1} \sum_{i=2}^{T} R_{i,i} - R^\textrm{ind}_{i}.
\end{align*}

FROMP achieves $6.1\pm0.7\%$, a much higher value compared to $0.17\pm0.9\%$ obtained with EWC and $1.8\pm3.1\%$ with VCL+coresets.
For backward transfer, we used the BWT metric defined in \citep{lopez2017gradient} which roughly captures the difference in accuracy obtained when a task is first trained and its accuracy after the final task. Again, higher is better and quantifies the gain obtained with the future tasks.
Here, FROMP has a score of $-2.6\pm0.9\%$, which is comparable to EWC's score of $-2.3\pm1.4\%$ but better than VCL+coresets which obtains $-9.2\pm1.8\%$. 

\begin{table*}[h]
    \centering
    \caption{Summary of metrics on Split CIFAR. FROMP outperforms the baselines EWC and VCL+coresets. All methods are run five times, with mean and standard deviation reported.
    }
    \begin{tabular}{llll}
        \toprule 
        \textbf{Method} & \textbf{Final average accuracy} & \textbf{Forward transfer} & \textbf{Backward transfer}\\
        \hline
            EWC             & $71.6 \pm 0.9\%$  & $0.17\pm0.9\%$  & $\mathbf{-2.3}\pm1.4\%$ \\
            VCL+coresets    & $67.4 \pm 1.4\%$  & $1.8\pm3.1\%$   &$-9.2\pm1.8\%$  \\
            FROMP           & $\mathbf{76.2} \pm 0.4\%$  & $\mathbf{6.1}\pm0.7\%$   & $\mathbf{-2.6}\pm0.9\%$ \\
        \bottomrule
    \end{tabular}
    \label{tab:cifar_metrics}
\end{table*}

%% file: sections/appendix/hyperparameters.tex
\section{Further details on experiments}
\label{app:hypers}

\subsection{Permuted MNIST}
We use the Adam optimiser \citep{adam} with Adam learning rate set to 0.001 and parameter $\beta_1=0.99$, and also employ gradient clipping. The minibatch size is 128, and we learn each task for 10 epochs. We use $\tau=0.5N$ for all algorithms, with 200 memorable points: FROMP, FRORP, FROMP-$L_2$ and FRORP-$L_2$. We use a fully connected single-head network with two hidden layers, each consisting of 100 hidden units with ReLU activation functions. We report performance after 10 tasks. 

\subsection{Split MNIST}
\label{app:split_mnist}
We use the Adam optimiser \citep{adam} with Adam learning rate set to 0.0001 and parameter $\beta_1=0.99$, and also employ gradient clipping. The minibatch size is 128, and we learn each task for 15 epochs. We use $\tau=10N$ for all algorithms, with 40 memorable points: FROMP, FRORP, FROMP-$L_2$ and FRORP-$L_2$. We use a fully connected multi-head network with two hidden layers, each with 256 hidden units and ReLU activation functions. 

\textbf{Smaller network architecture from \citet{swaroop2019improving}}. \citet{swaroop2019improving} use a smaller network than the network we use for the results in \cref{tab:mnist}. They train VCL on a single-hidden layer network with 100 hidden units (and ReLU activation functions). To ensure faithful comparison, we reran FROMP (with 40 memorable points per task) on this smaller network, obtaining a mean and standard deviation over 5 runs of $(99.2\pm0.1)\%$. This is an improvement from \cref{tab:mnist}, which uses a larger network. We believe this is due to the pruning effect described in \citet{swaroop2019improving}.

\textbf{Sensitivity to the value of $\tau$}. We tested FROMP and FROMP-$L_2$ with different values of the hyperparameter $\tau$. We found that $\tau$ can change by an order of magnitude without significantly affecting final average accuracy. Larger changes in $\tau$ led to greater than 0.1\% loss in accuracy.

\subsection{Split CIFAR}
\label{app:split_cifar}
We use the Adam optimiser \citep{adam} with Adam learning rate set to 0.001 and parameter $\beta_1=0.99$, and also employ gradient clipping. The minibatch size is 256, and we learn each task for 80 epochs. We use $\tau=10N$ for all algorithms, with 200 memorable points: FROMP, FRORP, FROMP-$L_2$ and FRORP-$L_2$.

\textbf{Numerical results on Split CIFAR}. We run all methods 5 times and report the mean and standard error. For baselines, we train from scratch on each task and jointly on all tasks achieving $(73.6 \pm 0.4)\%$ and $(78.1 \pm 0.3)\%$, respectively. The final average validation accuracy of FROMP is $(76.2 \pm 0.4)\%$, FROMP-$L_2$ is $(75.6 \pm 0.4)\%$, SI is $(73.5 \pm 0.5)\%$ (result from \citet{si}), EWC is $(71.6 \pm 0.9)\%$, VCL + random coreset is $(67.4 \pm 1.4)\%$. 

\begin{figure}[t]
 \centering
    \includegraphics[width=1.3in]{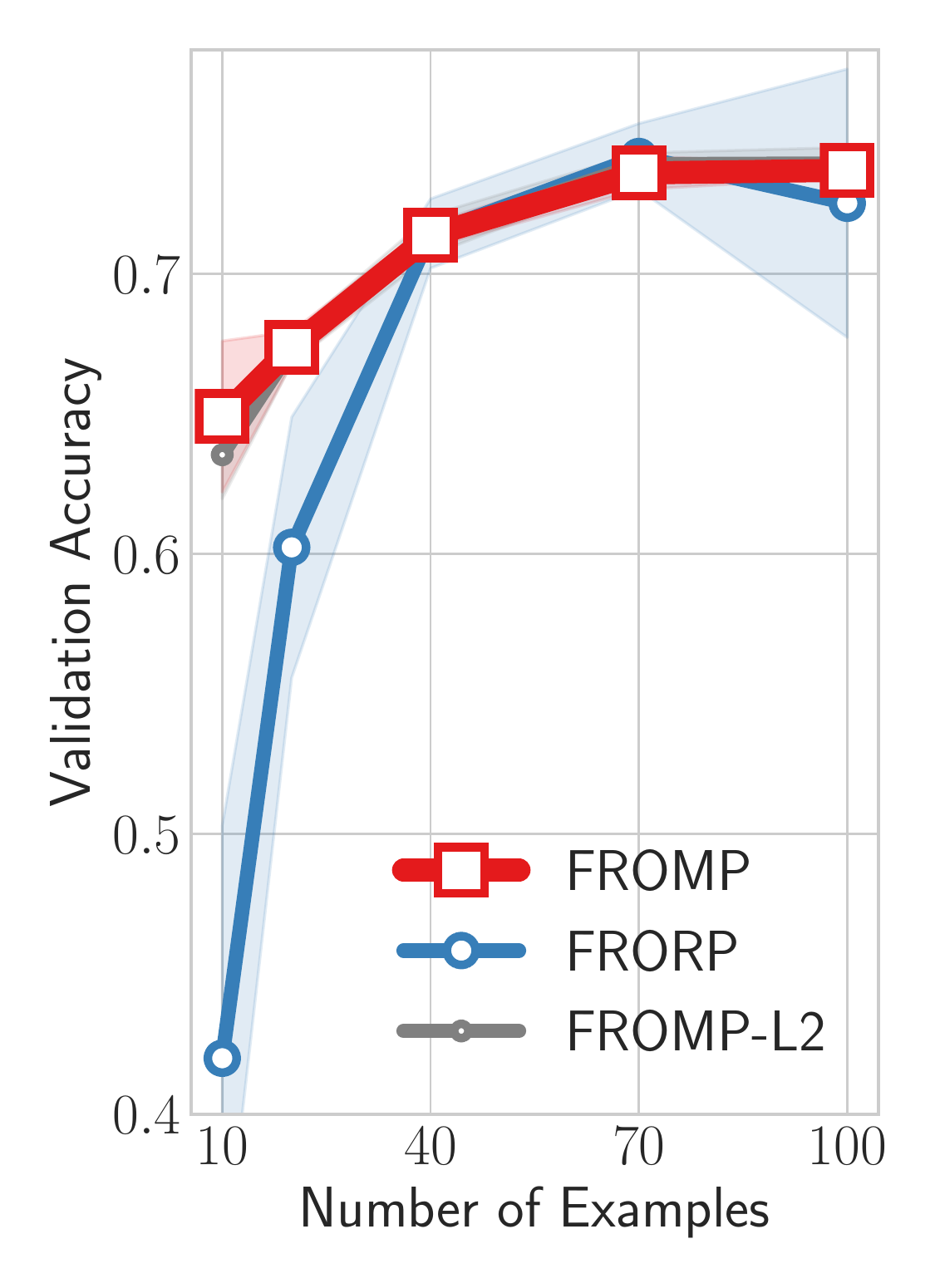}
    \caption{
    Results on Split CIFAR with 11 tasks as the number of memorable examples changes.
    A careful selection of memorable examples in FROMP gives better (/more consistent) results than random examples in FRORP, especially when the memory size is small.
}
    \label{fig:cifar11tasks_mem_size}
\end{figure}

\textbf{Longer task sequence: 11 tasks of Split CIFAR.} We also run Split CIFAR for 11 tasks instead of the standard 6 tasks, and compare FROMP with FROMP-$L_2$ and FRORP for different sizes of memorable past (\cref{fig:cifar11tasks_mem_size}). We find similar results to \cref{fig:mem_vs_acc_splitcifar} in the main text, with FROMP typically out-performing FRORP, especially at smaller memorable sizes, but being similar to FROMP-$L_2$.

\subsection{Fewer memorable past examples}
When we have fewer memorable past examples (for \cref{fig:permuted_mem_vs_acc,fig:mem_vs_acc_splitcifar}), we increase $\tau$ to compensate for the fewer datapoints. For example, for Split CIFAR, when we have 40 memorable past examples per task (instead of 200), we use $\tau=(200/40)*10N=50N$ (instead of $\tau=10N$ for 200 memorable past points).

%% file: sections/appendix/toydata_app.tex
\section{Toy data experiments}
\label{app:toydata}

\begin{figure*}[t]
    \centering
    \includegraphics[height=3cm]{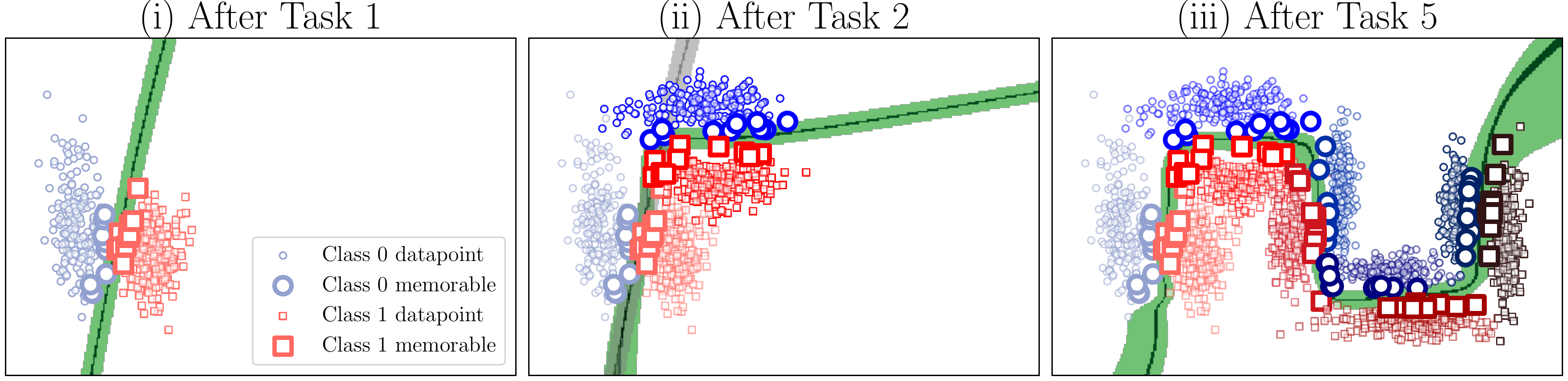}
    \caption{This figure demonstrates our approach on a toy dataset. Figure (i) shows the result of training on the first task where memorable past examples are shown with big markers. These points usually are the ones that support the decision boundary. 
    Figure (ii) shows the result after training on the second task where we see that the new network outputs are forced to give the same prediction on memorable past examples as the previous network. 
    The new decision boundary classifies both task 1 and 2 well.
    Figure (iii) shows the result after training on five tasks, along with the memorable-past of each task. With our method, the performance over past tasks is maintained.}
    \label{fig:app_illustration}
\end{figure*}

In this section, we use a 2D binary classification toy dataset with a small multi-layer perceptron to (i) demonstrate the brittleness and inconsistent behaviour of weight-regularisation, (ii) test FROMP's performance on different toy datasets of varying difficulty. 
As shown in \cref{fig:app_illustration} in \cref{app:toydata}, we find that weight-regularisation methods like VCL (+coresets) perform much worse than functional-regularisation, with lower accuracy, higher variance over random seeds, and visually bad decision boundaries.

The toy dataset we use is shown in \cref{fig:app_illustration}, along with how FROMP does well. In \cref{app:toy_vcl}, we show weight-space regularisation's inconsistent behaviour on this dataset, with results and visualisations. In \cref{app:dataset_var}, we show that FROMP performs consistently across many variations of the dataset. Finally, hyperparameters for our experiments are presented in \cref{app:toy_hypers}. For all these experiments, we use a 2-hidden layer single-head MLP with 20 hidden units in each layer.

\subsection{Weight-space regularisation's inconsistent behaviour}
\label{app:toy_vcl}

\begin{table*}[h]
    \centering
    \caption{Train accuracy of FROMP, VCL (no coresets), VCL+coresets and batch-trained Adam (an upper bound on performance) on a toy 2D binary classification dataset, with mean and standard deviations over 5 runs for VCL and batch Adam, and 10 runs for FROMP. `VCL' is without coresets. VCL-RP and FRORP have the same (random) coreset selections. VCL-MP is provided with `ideal' coreset points as chosen by an independent run of FROMP. VCL (no coreset) does very poorly, forgetting previous tasks. VCL+coresets is brittle with high standard deviations, while FROMP is stable.}
    \begin{tabular}{cccccc}
        \toprule 
        \textbf{FROMP} & \textbf{FRORP} & \textbf{VCL-RP} & \textbf{VCL-MP} & \textbf{VCL} & \textbf{Batch Adam}\\
        \hline
        $99.6 \pm 0.2\%$ & $98.5 \pm 0.6\%$ & $92 \pm 10\%$ & $85 \pm 14\%$ & $68 \pm 8\%$ & $99.70 \pm 0.03\%$ \\
        \bottomrule
    \end{tabular}
    \label{tab:vcl_toy}
\end{table*}

\cref{tab:vcl_toy} summarises the performance (measured by train accuracy) of FROMP and VCL+coresets on a toy dataset similar to that in \cref{fig:app_illustration}. FROMP is very consistent, while VCL (with coresets) is extremely brittle: it can perform well sometimes (1 run out of 5), but usually does not (4 runs out of 5). This is regardless of the coreset points chosen for VCL. Note that coresets are chosen independently of training in VCL. Without coresets, VCL forgets many past tasks, with very low performance. 

\begin{figure*}[h!]
    \centering
    \begin{subfigure}[b]{0.3\textwidth}
        \includegraphics[width=\textwidth]{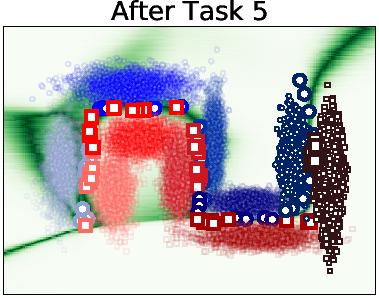}
    \end{subfigure}
    \begin{subfigure}[b]{0.3\textwidth}
        \includegraphics[width=\textwidth]{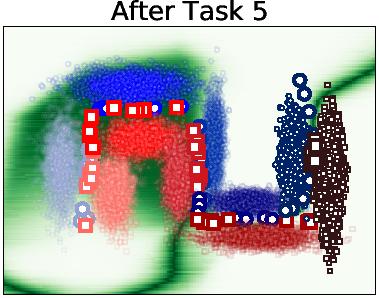}
    \end{subfigure}
    \begin{subfigure}[b]{0.3\textwidth}
        \includegraphics[width=\textwidth]{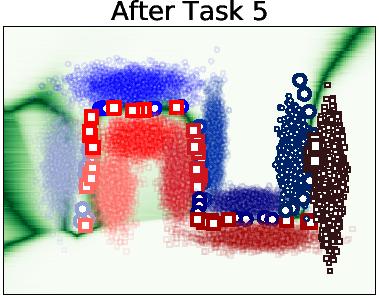}
    \end{subfigure}
    \caption{Three runs of VCL-MP on toy 2D data. These are the middle performing 3 runs out of 5 runs with different random seeds. VCL's inconsistent behaviour is clear.}
    \label{fig:VCL_toy_runs}
\end{figure*}

For VCL-MP, the coreset is chosen as the memorable past from an independent run of FROMP, with datapoints all on the task boundary. This selection of coreset is intuitively better than a random coreset selection. The results we show here are not specific to coreset selection. Any coreset selection (whether random or otherwise) all show the same inconsistency when VCL is trained with them. We provide visualisations of the brittleness of VCL in \cref{fig:VCL_toy_runs}.

\subsection{Dataset variations}
\label{app:dataset_var}

\begin{figure*}[h]
    \centering
    \begin{subfigure}[b]{0.45\textwidth}
        \includegraphics[width=\textwidth]{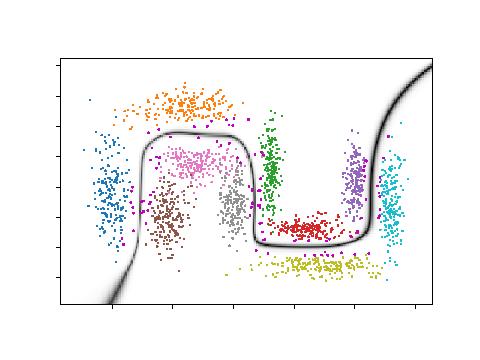}
    \end{subfigure}
    \begin{subfigure}[b]{0.45\textwidth}
        \includegraphics[width=\textwidth]{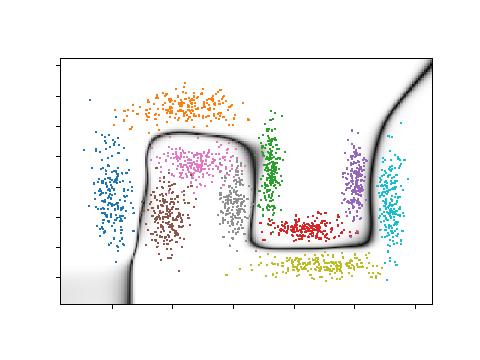}
    \end{subfigure}
    \caption{FROMP (middle performing of 5 runs) and batch Adam on a dataset 10x smaller (400 points per task).}
    \label{fig:dataset_0_200}
\end{figure*}

\begin{figure*}[h]
    \centering
    \begin{subfigure}[b]{0.45\textwidth}
        \includegraphics[width=\textwidth]{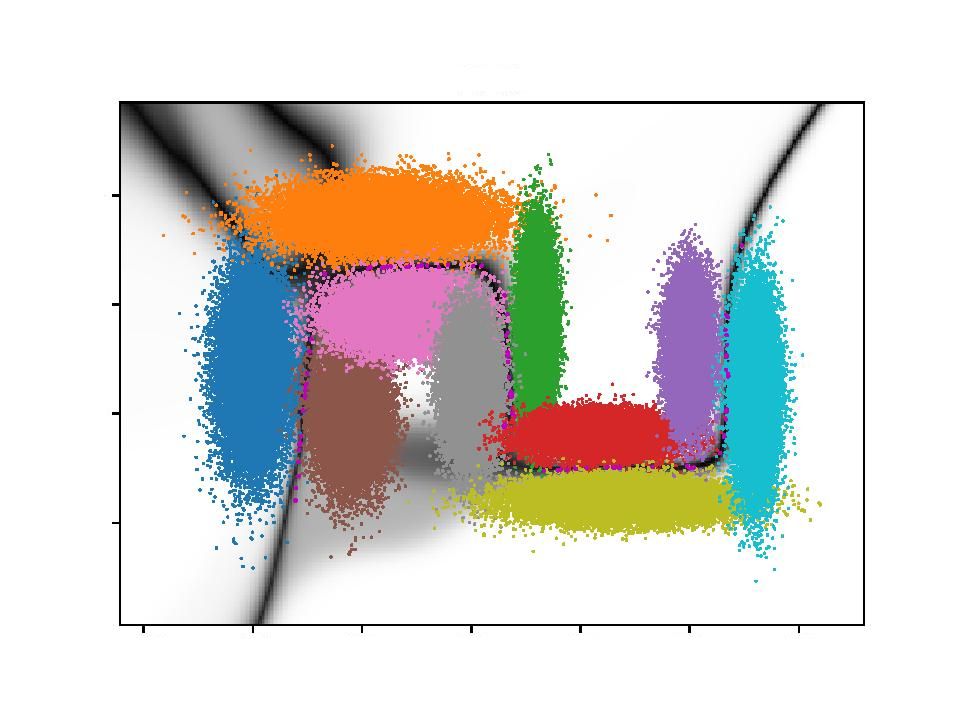}
    \end{subfigure}
    \begin{subfigure}[b]{0.45\textwidth}
        \includegraphics[width=\textwidth]{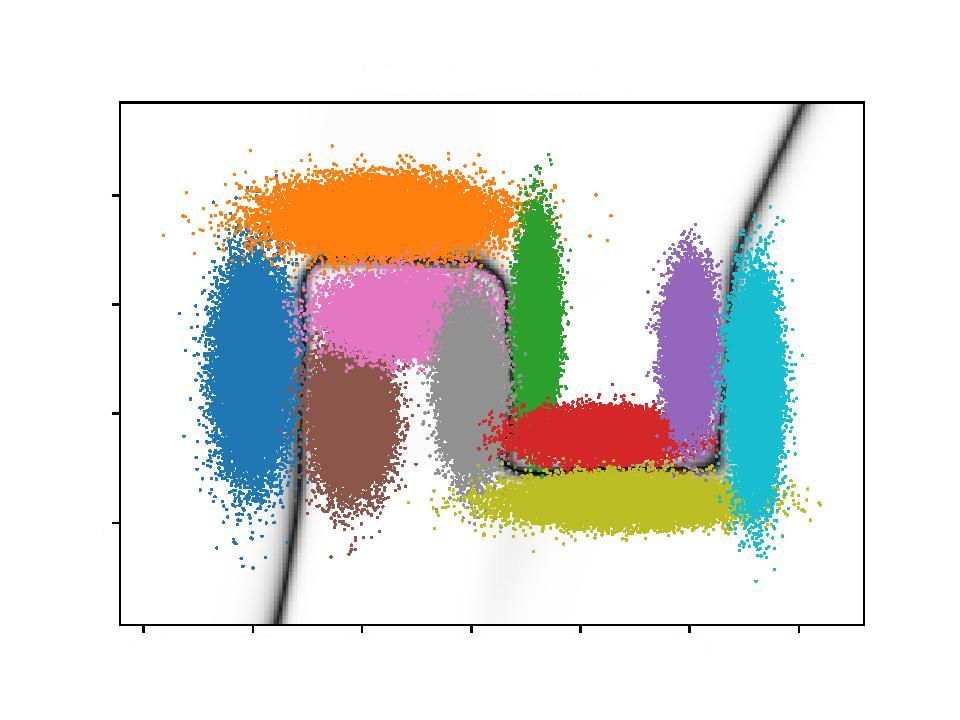}
    \end{subfigure}
    \caption{FROMP (middle performing of 5 runs), left, and batch Adam, right, on a dataset 10x larger (40,000 points per task).}
    \label{fig:dataset_0_20000}
\end{figure*}

\begin{figure*}[h]
    \centering
    \begin{subfigure}[b]{0.45\textwidth}
        \includegraphics[width=\textwidth]{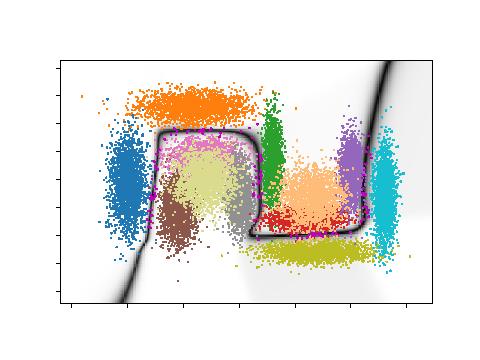}
    \end{subfigure}
    \begin{subfigure}[b]{0.45\textwidth}
        \includegraphics[width=\textwidth]{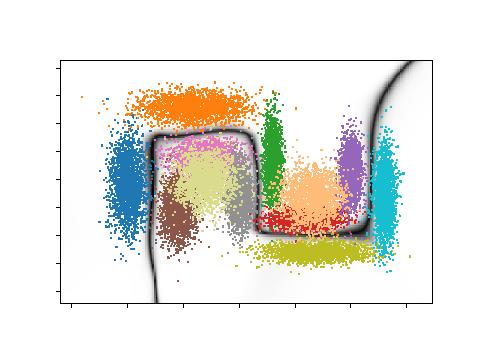}
    \end{subfigure}
    \caption{FROMP (middle performing of 5 runs), left, and batch Adam, right, on a dataset with a new, easy, 6th task.}
    \label{fig:dataset_1}
\end{figure*}

\begin{figure*}[h]
    \centering
    \begin{subfigure}[b]{0.45\textwidth}
        \includegraphics[width=\textwidth]{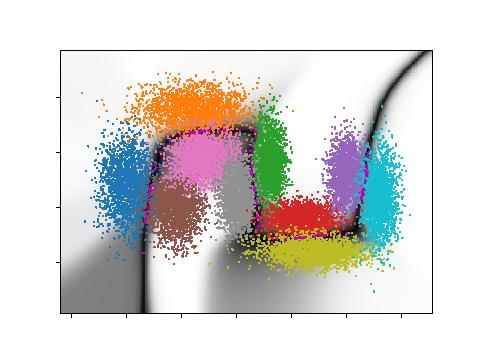}
    \end{subfigure}
    \begin{subfigure}[b]{0.45\textwidth}
        \includegraphics[width=\textwidth]{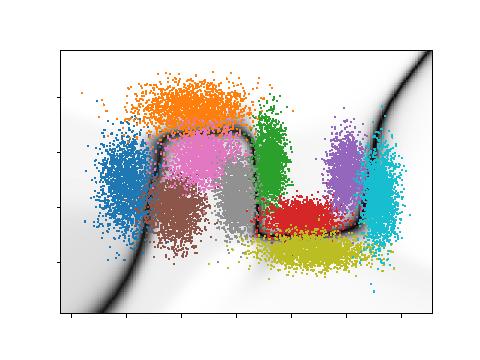}
    \end{subfigure}
    \caption{FROMP (middle performing of 5 runs), left, and batch Adam, right, on a dataset with increased standard deviations of each class' points, making classification tougher.}
    \label{fig:dataset_2}
\end{figure*}

\begin{figure*}[h]
    \centering
    \begin{subfigure}[b]{0.45\textwidth}
        \includegraphics[width=\textwidth]{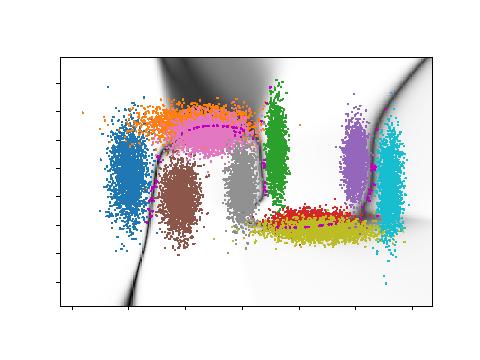}
    \end{subfigure}
    \begin{subfigure}[b]{0.45\textwidth}
        \includegraphics[width=\textwidth]{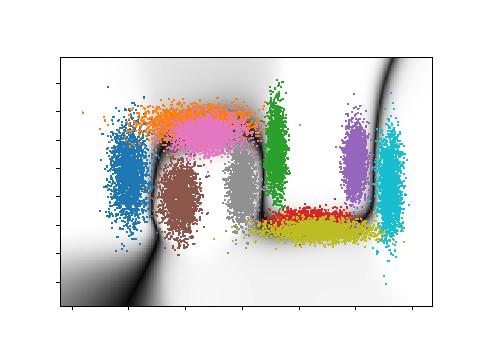}
    \end{subfigure}
    \caption{FROMP (middle performing of 5 runs), left, and batch Adam, right, on a dataset with 2 tasks having overlapping data, which is not separable.}
    \label{fig:dataset_3}
\end{figure*}

\crefrange{fig:dataset_0_200}{fig:dataset_3} visualise the different dataset variations presented in \cref{tab:toy_results}. We pick the middle performing FROMP run (out of 5) and batch Adam run to show.

\begin{table*}[ht]
    \centering
    \caption{Train accuracy of FROMP and batch-trained Adam (upper bound on performance) on variations of a toy 2D binary classification dataset, with mean and standard deviations over 10 runs (3 runs for Adam). FROMP performs well across variations. VCL (with coresets) performs significantly worse even on the original dataset ($92 \pm 10\%$). See \cref{app:dataset_var} for further experiments and for visualisations.}
    \begin{tabular}{lcc}
        \toprule 
        \textbf{Dataset variation} & \textbf{FROMP} & \textbf{Batch Adam}\\
        \hline
        Original dataset                                & $99.6 \pm 0.2\%$      & $99.7 \pm 0.0\%$ \\
        10x less data (400 per task)                    & $99.9 \pm 0.0\%$      & $99.7 \pm 0.2\%$ \\
        10x more data (40000 per task)                  & $96.9 \pm 3.0\%$      & $99.7 \pm 0.0\%$ \\
        Introduced 6th task                             & $97.8 \pm 3.3\%$      & $99.6 \pm 0.1\%$ \\
        Increased std dev of each class distribution    & $96.0 \pm 2.4\%$      & $96.9 \pm 0.4\%$ \\
        2 tasks have overlapping data                   & $90.1 \pm 0.8\%$      & $91.1 \pm 0.3\%$ \\
        \bottomrule
    \end{tabular}
    \label{tab:toy_results}
\end{table*}

\subsection{VCL and FROMP hyperparameter settings for toy datasets}
\label{app:toy_hypers}

\textbf{FROMP.} We optimised the number of epochs, Adam learning rate, and batch size. We optimised by running different hyperparameter settings for 5 runs on the toy dataset in \cref{fig:app_illustration}, and picking the settings with largest mean train accuracy. We found the best settings were: number of epochs=50, batch size=20, learning rate=0.01. The hyperparameters were then fixed across all toy data experimental runs, including across dataset variations (number of epochs was appropriately scaled by 10 if dataset size was scaled by 10).

\textbf{VCL+coresets.} We optimised the number of epochs, the number of coreset epochs (because VCL+coresets trains on non-coreset data first, then on coreset data just before test-time: see \citet{vcl}), learning rate (we use Adam to optimise the means and standard deviations of each parameter), batch size, and prior variance. We optimised by running various settings for 5 runs and picking the settings with largest mean train accuracy. We found the best settings were: number of epochs=200, number of coreset epochs=200, a standard normal prior (variance=1), batch size=40, learning rate=0.01. VCL is slow to run (an order of magnitude longer) compared to the other methods (FROMP and batch Adam).

\subsection{Importance of kernel being over all layer weights}
\label{app:importance_all_layers}

\begin{figure*}[h]
    \centering
    \begin{subfigure}[b]{\textwidth}
        \includegraphics[width=\textwidth]{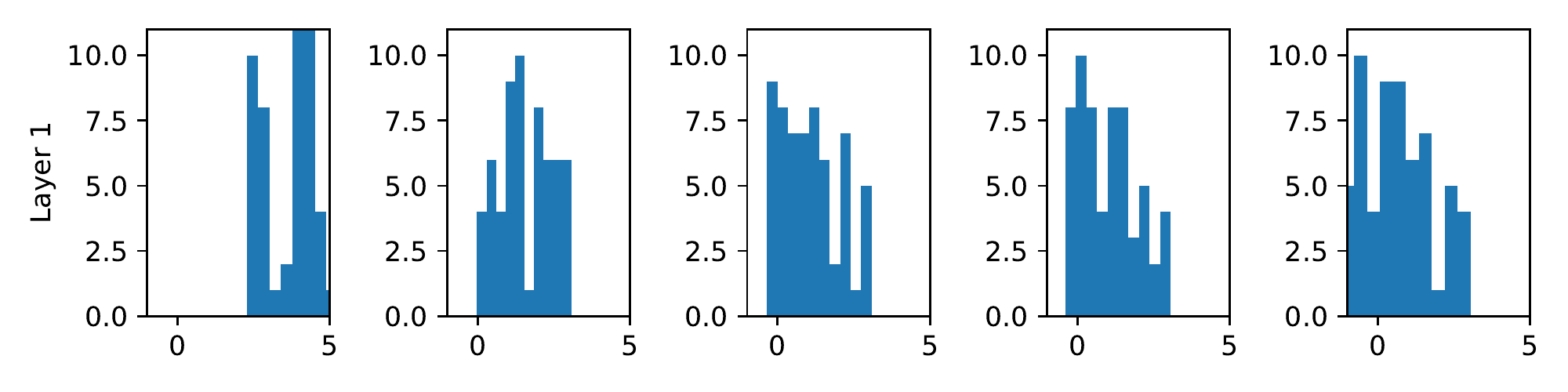}
    \end{subfigure}
    \begin{subfigure}[b]{\textwidth}
        \includegraphics[width=\textwidth]{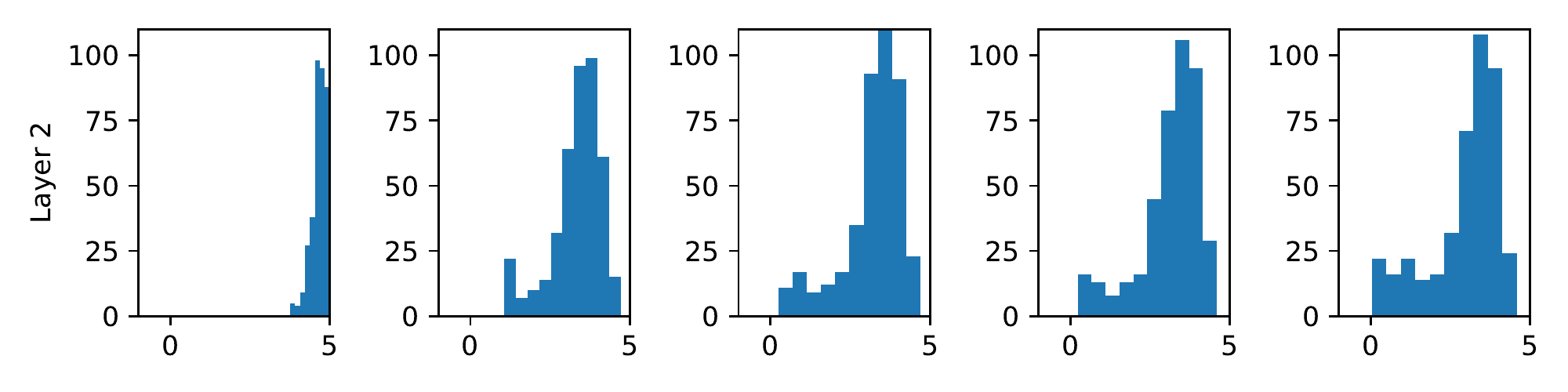}
    \end{subfigure}
        \begin{subfigure}[b]{\textwidth}
        \includegraphics[width=\textwidth]{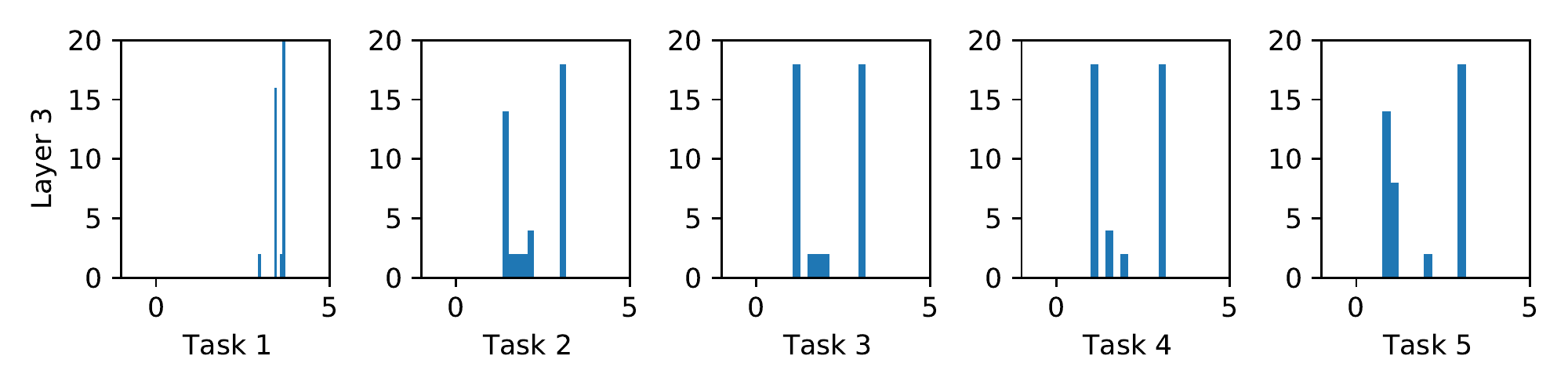}
    \end{subfigure}
    \caption{Histogram of entropy of distribution the distribution of weights for different layers (row) and task (columns). For each layer, we take all weights and plot the histogram of their entropies. Left-most is after the first task, and right-most is after the last task. We see that the entopy is high across layers, implying that there is significant uncertainties about the weights for all of them, not only the last layer (layer 3 in this case).}
    \label{fig:layerwise_entropies}
\end{figure*}

In this section, we show the importance of using all weights of the neural network, instead of just the last layer. Our kernel is over all weights from all layers. We run the same toy experiment, and consider the entropies of the Gaussian distributions for weights in each layer. We plot the histogram of these entropies in \cref{fig:layerwise_entropies}. As can be seen, all layers have weights with high uncertainty (high entropy), especially for the first few tasks. Note that as we train for more tasks, we expect the uncertainties to reduce as our network parameters become more certain having seen more data.

Therefore, by considering uncertainties across weights in all layers, instead of just the last layer, we might expect better performance.

%% file: sections/appendix/task_boundary_detection.tex
\section{Task boundary detection}
\label{app:task_boundary_detection}

In this section, we consider the case where data is separated into tasks, but we are not provided task boundaries during training. Our goal is to detect the task boundaries. Many of the ideas in this section are inspired from \citet{titsias2019functional} Section~3.

We consider 10 tasks of Permuted MNIST, with minibatches arriving without task ID information. We wish to automatically detect when a new minibatch belongs to a new task. We use the same network and hyperparameters as in \cref{app:hypers}. 

The key insight is that, when we first see data from a new task, we expect this data to be far (in input space) from data we have observed so far. Therefore, predictions over this new data with our current network parameters, $\vm_t$, should be similar to predictions with our prior network parameters, $\vm_{t-1}$. This is in contrast to when we see data from the current task, when predictions with our current network parameters will be very different to our prior network parameters.

Using this insight, we perform a test on every new minibatch of data, in order to determine whether it is from a new task or not. This test is performed before training on the minibatch. 

For every new minibatch, we:
\begin{enumerate}
    \item Calculate $(m_{t,i} - m_{t-1,i})^2$ for each sample $i$ in the minibatch where $m_{t,i}$ and $m_{t-1,i}$ are predictive mean obtained using current and past networks respectively. When we see a new task, we expect this value to be small.
    \item Calculate Welch's t-test statistic between the current and the previous minibatch's samples. For the multi-class setup of Permuted MNIST, we repeat this for each function, and average this statistic across the functions.
    \item If the statistic is sufficiently high (above a threshold), we detect a new task. 
\end{enumerate}

We find that this method is very good at determining task boundaries. We always successfully recognise a task change, with no mistakes, over a wide range of thresholds. Note that we do not conduct the test for the first 10 iterations of training on a new task. 

\begin{figure*}[h]
 \centering
  \begin{subfigure}[b]{0.45\textwidth}
     \centering
     \includegraphics[width=\textwidth]{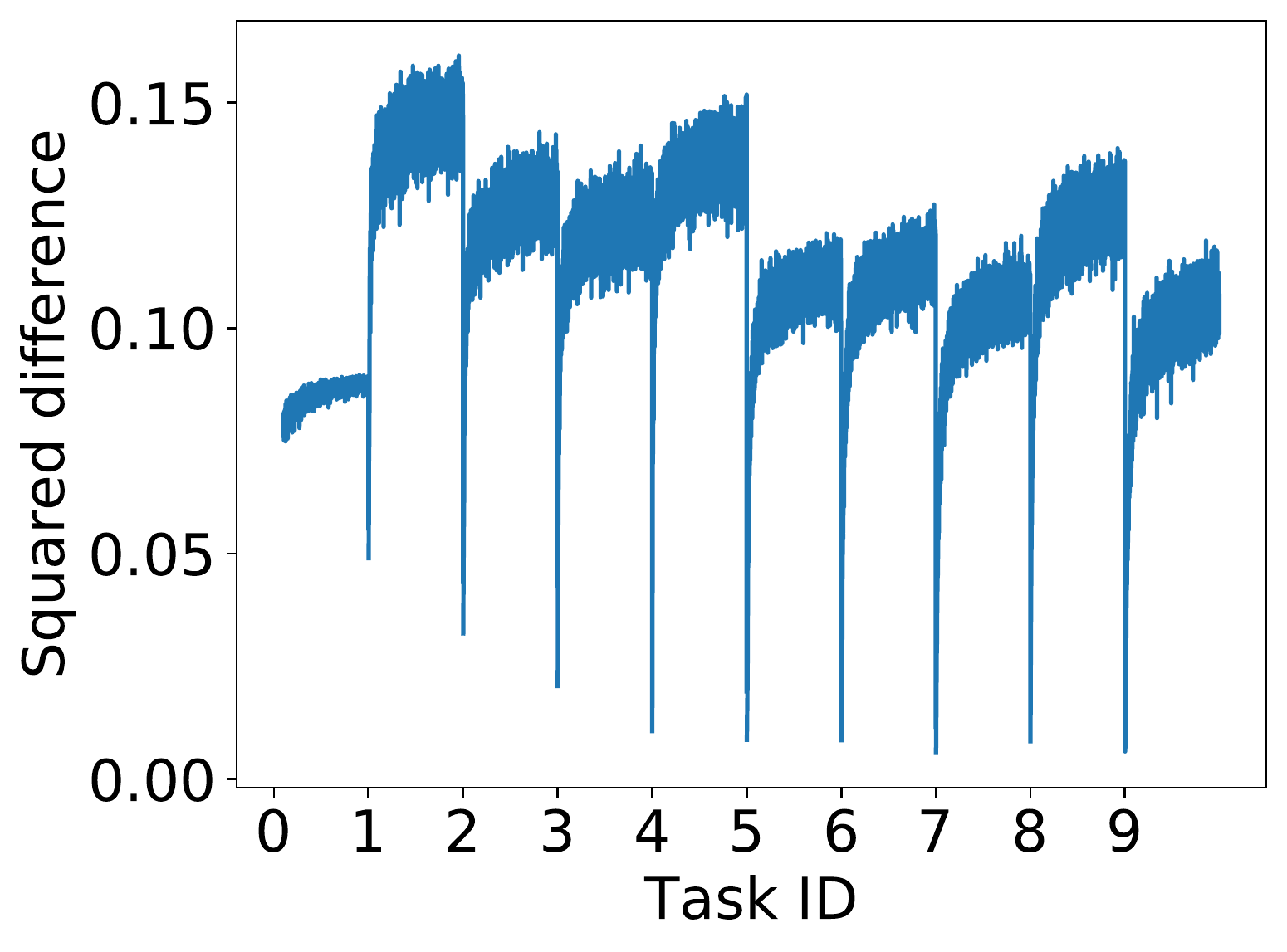}
     \caption{Square of difference in mean predictions}
     \label{fig:taskdetection_squareddiff}
 \end{subfigure}
 \hfill
 \begin{subfigure}[b]{0.42\textwidth}
     \centering
     \includegraphics[width=\textwidth]{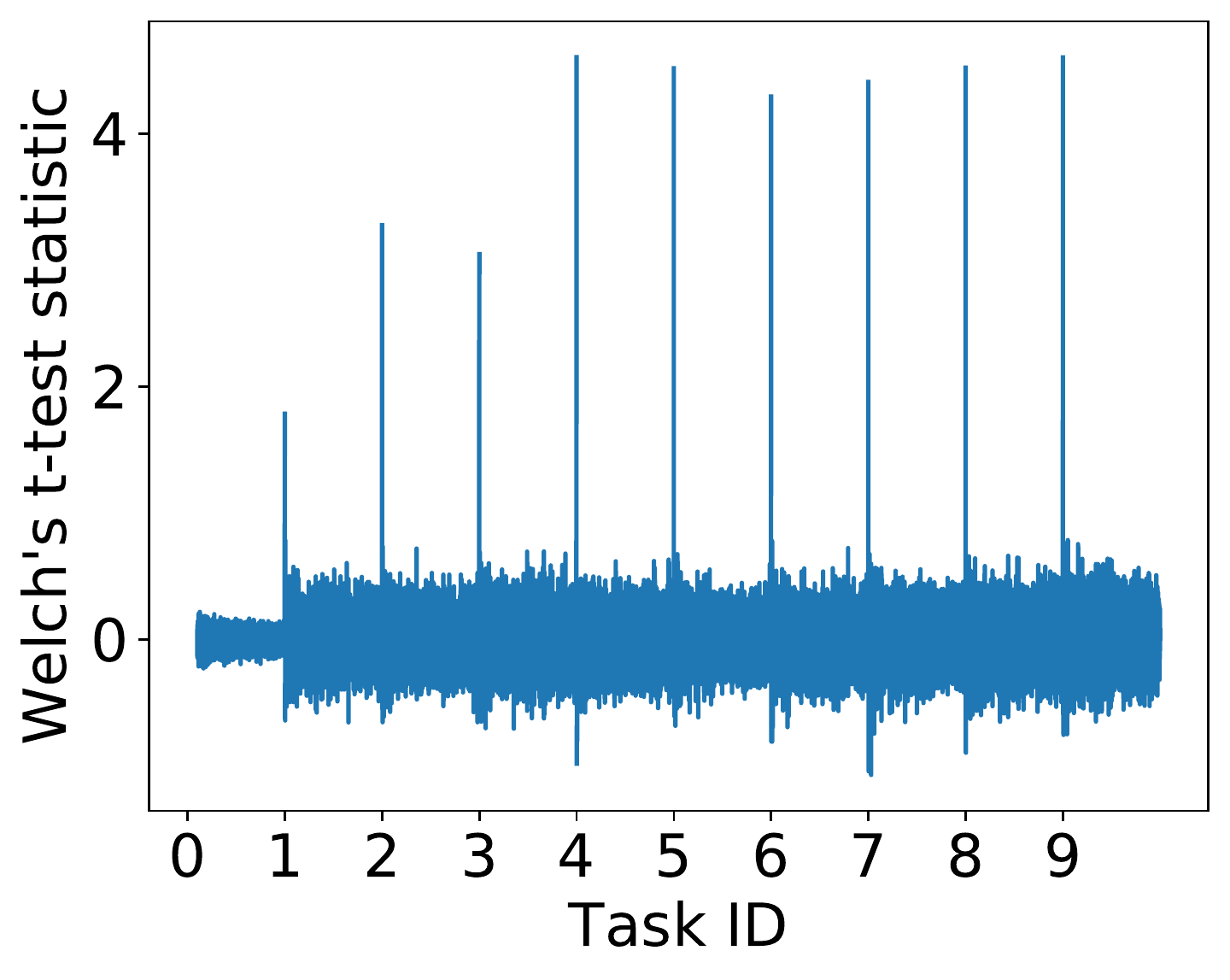}
     \caption{Welch's t-test statistic}
     \label{fig:taskdetection_test}
 \end{subfigure}
    \caption{Detecting task boundary changes in 10 tasks of Permuted MNIST.
    (a) The square of the difference in means reduces noticeably whenever a minibatch from a new task is seen for the first time.
    (b) We can perform Welch's t-test to detect these changes, and threshold on this value to detect a new task.
    }
    \label{fig:taskdetection}
\end{figure*}

We plot the Welch's t-test statistic between minibatches in \cref{fig:taskdetection} for a specific run. As can be seen, we can use a range of threshold values (approximately 0.9 to 1.8, limited by detecting the very first task change) to successfully recognise that the task has changed. 

We found that using just mean predictions to be good enough for determining task boundaries in this setting. Ideally, in more complicated scenarios, we might want to use the full GP predictive distribution, and compare that to the predictive distribution from the GP prior. We could then use a divergence to determine how similar the two distributions are, with the expectation that new tasks have small divergence.

%% file: sections/appendix/camera_ready_changes.tex
\section{Changes in the camera-ready version compared to the submitted version}

\begin{figure}[t]
 \centering
     \begin{subfigure}{1.2in}
         \centering
         \includegraphics[width=1.2in]{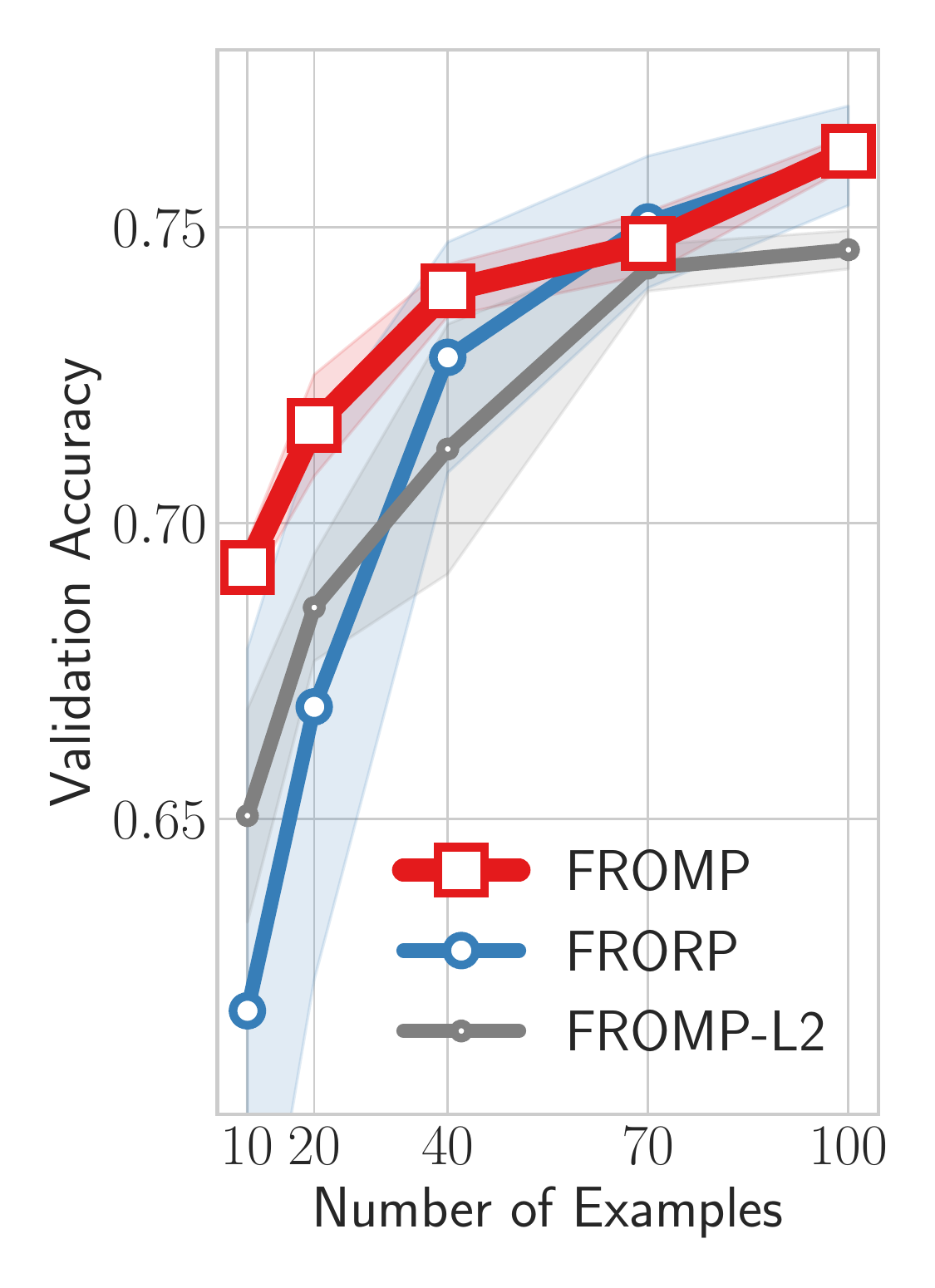}
         \caption{Split CIFAR}
         \label{fig:OLD_mem_vs_acc_splitcifar}
     \end{subfigure}
     \begin{subfigure}{1.2in}
        \centering
        \includegraphics[width=1.2in]{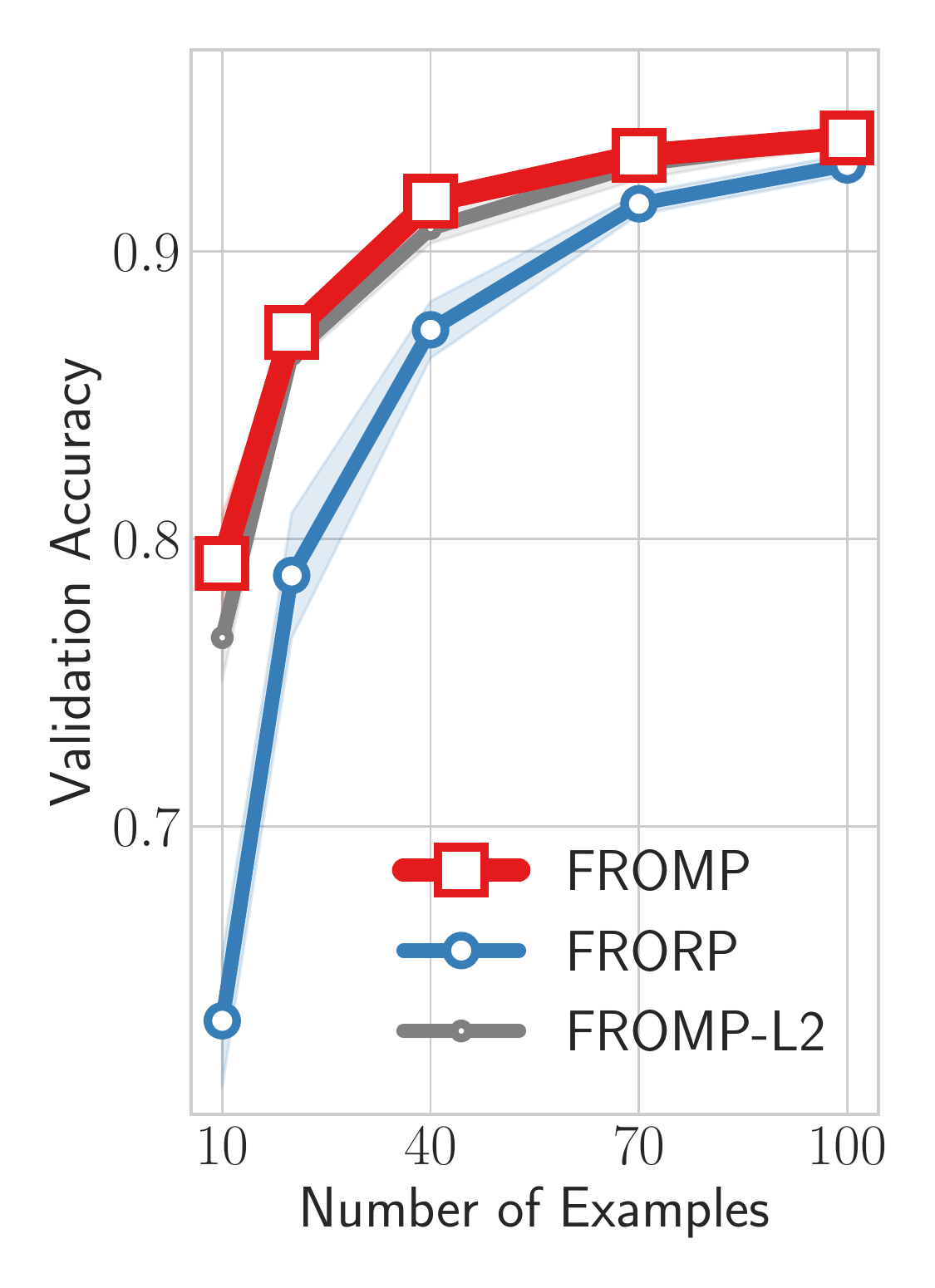}
        \caption{Permuted MNIST}
        \label{fig:OLD_permuted_mem_vs_acc}
     \end{subfigure}
    \caption{
    Previous figures for average accuracy with respect to the number of memorable examples. 
}
    \label{fig:OLD_past_samples}
\end{figure}

\begin{itemize}
    \item We expanded the Related Works section.
    \item We added the task boundary detection experiment (\cref{app:task_boundary_detection}).
    \item We ran Split CIFAR on 11 tasks (\cref{app:hypers}).
    \item Added \cref{app:importance_all_layers} discussing (with a toy visualisation) the importance of using all layer weights in the kernel matrix, not just the last layer weights.
    \item We updated \cref{fig:mem_vs_acc_splitcifar} and \cref{fig:permuted_mem_vs_acc} in the main text using the newest hyperparameters we found. The new plot shows that FROMP and FROMP-$L_2$ are similar in \cref{fig:mem_vs_acc_splitcifar}, but slightly further apart in \cref{fig:permuted_mem_vs_acc}. The old figures are in \cref{fig:OLD_past_samples}, which we believe should be attainable with different hyperparameters.
\end{itemize}

%% file: sections/appendix/author_contri.tex
\section{Author Contributions Statement}
List of Authors: Pingbo Pan, Siddharth Swaroop, Alexander Immer, Runa Eschenhagen, Richard E. Turner, Mohammad Emtiyaz Khan.

P.P, S.S., and M.E.K. conceived the original idea of using DNN2GP for continual learning. This was then discussed with R.E., R.T., and A.I. 
The DNN2GP result from Section 3.1 is due to A.I. The memorable past method in Section 3.2 is due to M.E.K. 
The FROMP algorithm in Algorithm 1 was originally conceived by P.P., S.S. and M.E.K. The idea of functional prior was conceived by S.S. and M.E.K. Based on this idea, S.S. and M.E.K. wrote a derivation using the variational approach, which is currently written in Section 3.3. 
R.E., A.I. and R.T. regularly provided feedback for the main methods.

P.P. conducted all experiments, with feedback from M.E.K., A.I., R.E, and S.S.
S.S. made corrections to some of the code, fixed hyperparameter reporting, and also did baseline comparisons.

The first version of the paper was written by M.E.K. with some help from the other authors. S.S revised the paper many times and also rewrote many new parts. Detailed derivation in Appendix is written by S.S. and M.E.K. The authors A.I., R.E. and R.T. provided feedback during the writing of the paper.

M.E.K. and S.S. led the project.